\documentclass[10pt,twocolumn,letterpaper]{article}

\usepackage[pagenumbers]{cvpr}      %

\usepackage{multirow}
\usepackage{tikz}
\usetikzlibrary{calc}
\usepackage{pgfplots}
\usepgfplotslibrary{groupplots}
\pgfplotsset{compat=1.17}
\definecolor{wvblue}{HTML}{2F5BD3}
\definecolor{wvorange}{HTML}{D9822B}
\definecolor{wvgreen}{HTML}{2E8B57}
\usepackage{stfloats} %

\definecolor{cvprblue}{rgb}{0.21,0.49,0.74}
\usepackage[pagebackref,breaklinks,colorlinks,allcolors=cvprblue]{hyperref}

\def\confName{CVPR}
\def\confYear{2027}
\DeclareRobustCommand{\dtcwpt}{\textup{DT-$\mathbb{C}$WPT}}

\title{WaveView: The Novel View Is in the Noise}

\author{
Kaiwen Wang$^{1}$ \quad Jan-Hendrik Pauls$^{1,2}$ \quad Yinzhe Shen$^{1}$ \quad Frank Bieder$^{3}$ \quad Christoph Stiller$^{1}$
\\
{$^{1}$Karlsruhe Institute of Technology, Germany} \quad
{$^{2}$Parallel Domain, Germany}\\
{$^{3}$FZI Research Center for Information Technology, Germany}\\
{\tt\small kaiwen.wang@kit.edu}
}

\begin{document}
\twocolumn[{%
\renewcommand\twocolumn[1][]{#1}%
\maketitle
\vspace{-2em}
\begin{center}
  \includegraphics[width=\textwidth]{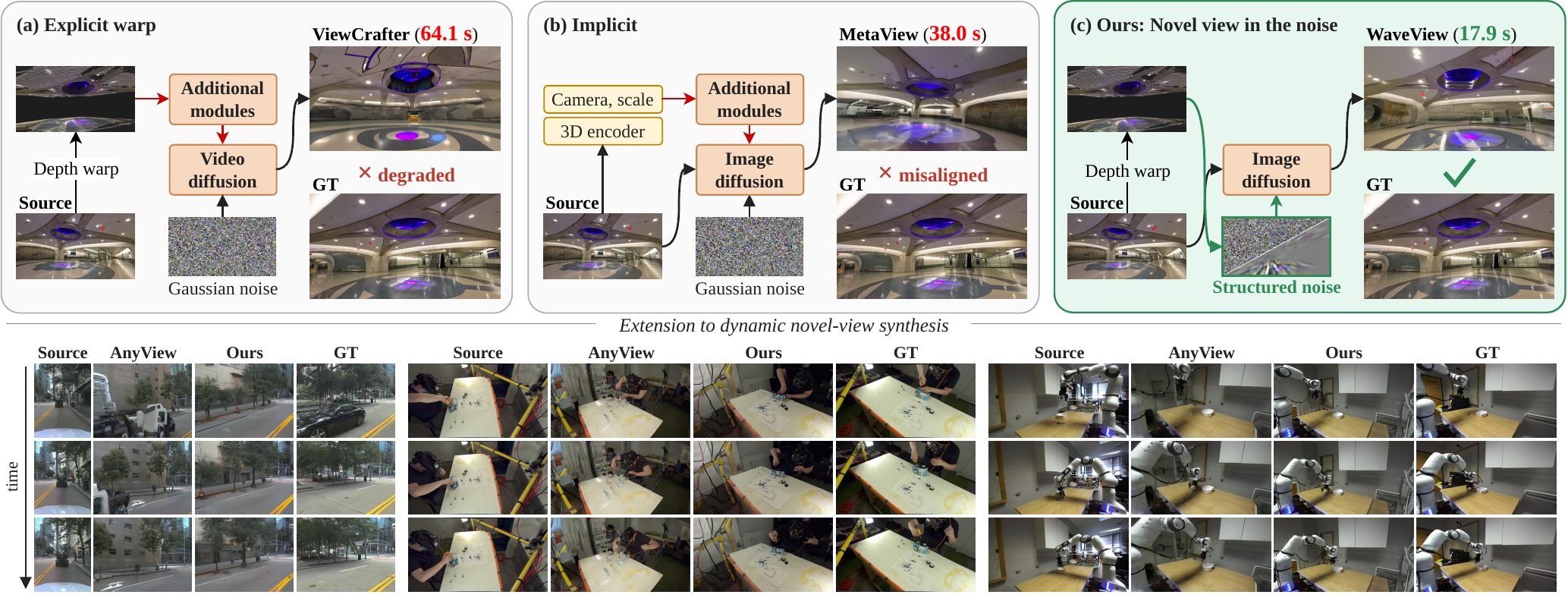}
  \captionof{figure}{\textbf{Comparison of generative novel view synthesis (NVS) methods.} (a) Explicit warp methods guide video diffusion models with the reprojection. (b) Implicit methods encode the camera parameters or 3D cues. Both introduce additional modules. (c) We introduce \textit{WaveView}, which anisotropically injects the geometric prior of the novel view into the phase of the initial noise without modifying the architecture. Inference times are given in parentheses. We further extend WaveView to dynamic NVS, where its output stays closer to the ground-truth layout than AnyView. }
  \label{fig:teaser}
\end{center}%
}]
\begin{abstract}
Generative novel view synthesis (NVS) employs a diffusion model to render a static or dynamic scene from a target camera view.
Explicit warp methods complete a reprojection of the source with video diffusion models, and implicit methods encode the camera parameters or 3D cues.
Both add inference overhead, and the generated layout often drifts from the ground truth.
We present \textbf{WaveView}, which injects the warped source into the phase of the initial noise of a pretrained generator.
The Dual-Tree Complex Wavelet Packet Transform (\dtcwpt) allows local and anisotropic phase injection.
The generator is thus constrained less in disoccluded regions and along the epipolar line. 
WaveView keeps the architecture and adds negligible overhead.
On the DL3DV benchmark, it outperforms all baselines in every overlap bin. 
In the hardest bin, it gains 0.84\,dB PSNR on the covisible region over MetaView, which shares its generator, while running 2.1 times faster.
For dynamic NVS, fine-tuning AnyView with the same noise raises its zero-shot PSNR on AnyViewBench from 11.58 to 12.35\,dB.
\end{abstract}
    
\section{Introduction}
\label{sec:intro}

Novel view synthesis from a single image asks for a scene as seen from a camera view that was never observed.
When the camera moves far, little of the target view is visible in the source, and a generative model has to complete it.
Two requirements then hold at once: the originally visible content must land where the geometry places it, and the completed content must look plausible.

Explicit warp methods reproject the source with an estimated depth and let a video diffusion model complete the warp~\cite{yu2025viewcrafter,ren2025gen3c,huang2025voyager}.
The warp fixes the layout.
But the generator is not told which pixels of the warp can be trusted, so depth errors, stretched surfaces, and the ragged borders of disocclusions are carried into the output.
These methods also often generate a video to obtain a single view, making them 3.6 to 27 times slower than ours (see \cref{sec:runtime} for a comparison).

Implicit methods omit the warp and provide the camera, and in some cases implicit geometry features, as a pose or in the positional encoding of the source tokens~\cite{bai2025pefield,cai2026metaview}, relying on the learned distribution of the generator.
Their outputs are clean, but their layout drifts as the viewpoint change grows.
\cref{fig:teaser} shows both failures on one example: ViewCrafter~\cite{yu2025viewcrafter} distorts the geometry, and MetaView~\cite{cai2026metaview} places the hall off the layout that the warp could have provided.
For both, generation starts from Gaussian noise.

The noise does not have to be neutral, however.
Phase-preserving diffusion shows that a pretrained generator follows the structure carried by the phase of its initial noise, while the magnitude remains Gaussian~\cite{zeng2025neuralremaster,wang2026psipd}.
The noise can therefore guide the layout of a novel view.
Phase fixes where structure lies, while the random magnitude leaves the generator free to correct it.

In this paper, we go one step further and challenge isotropy.
We present \textbf{WaveView}, which injects a geometric prior of the novel view into the initial noise of a pretrained generator.
A depth warp places the source at the target view, and the noise takes its phase from this warp and its magnitude from a Gaussian sample.
The phase is not trusted uniformly: an error in depth moves a pixel only along its epipolar line.
We therefore release the phase along that line as the depth uncertainty grows and keep its information across the line.
An isotropic floor absorbs the error of the warp itself, and disoccluded regions remain pure noise.

As the generator for static NVS, we use Qwen-Image-Edit~\cite{wu2025qwenimage}, adapted to this noise with a LoRA and left unchanged in architecture.
In contrast to the additional modules of both families, estimating depth and constructing the noise add only 0.46\,s.
On DL3DV~\cite{ling2024dl3dv}, WaveView is best in every metric on the Easy, Medium, and Hard bins against explicit warp and implicit methods.
On Hard, it gains 0.84\,dB PSNR over MetaView, which shares its generator, on the region observed in the source.
Releasing the phase along the epipolar line outperforms the same release spread over every direction, and the gap grows with the depth uncertainty (\cref{fig:ablation}).
For dynamic NVS, fine-tuning AnyView~\cite{vanhoorick2026anyview} with the structured noise raises its average PSNR on the zero-shot split of AnyViewBench from 11.58 to 12.35\,dB, the best of all compared methods, with training data that follow the distribution of AnyView.

In summary, our contributions are:
\begin{itemize}
  \item We guide the novel view through a geometric prior in the phase of the initial noise rather than through additional modules.
  \item We inject the phase anisotropically by releasing it along the epipolar line according to the depth uncertainty and keeping it across the line. Disoccluded regions are left as pure Gaussian noise.
  \item WaveView outperforms explicit warp and implicit methods, including MetaView with the same generator, on all three bins of DL3DV while running faster.
  The same noise improves AnyView on the zero-shot split of AnyViewBench in every average metric.
\end{itemize}
\section{Related Work}
\label{sec:related}

\subsection{Novel View Synthesis}
\label{sec:related-nvs}

Novel view synthesis estimates scene appearance at an unobserved camera view, for a static scene or, for a dynamic scene, along a new camera trajectory.
Reconstruction methods represent the scene and render that camera.
For a static scene, per-scene optimization fits a radiance field or a set of 3D Gaussians to many calibrated views~\cite{mildenhall2020nerf,kerbl20233dgs}.
The fit does not transfer to a new scene.
Feed-forward models predict the representation from one or a few images.
pixelNeRF conditions a radiance field on pixel-aligned features and can render from a single view~\cite{yu2021pixelnerf}.
MVSNeRF and IBRNet aggregate a sparse set of posed views, the former through a plane-swept cost volume~\cite{chen2021mvsnerf,wang2021ibrnet}.
The same division appears for Gaussians.
Splatter Image maps one image to one Gaussian per pixel~\cite{szymanowicz2024splatter}.
pixelSplat and MVSplat predict Gaussians from image pairs or sparse views, and DepthSplat couples that prediction to a multi-view depth model~\cite{charatan2024pixelsplat,chen2024mvsplat,xu2024depthsplat}.
For a dynamic scene, 4D Gaussians extend the representation over time and are fitted per scene to a video~\cite{wu20244dgs}.
Reconstructing a dynamic scene from a monocular video requires geometry and motion to be recovered jointly~\cite{wang2025som}.
Diffusion-based novel view synthesis uses generative priors to synthesize unobserved content, which reconstruction cannot recover.

\subsection{Diffusion Models for Novel Views}
\label{sec:related-diffusion}

For static NVS, pose-conditioned diffusion maps a source image and a relative camera to the novel view~\cite{liu2023zero123,sargent2024zeronvs}.
Explicit warp methods pass a reprojection of the source to the generator, and ViewCrafter, GEN3C, and Voyager complete it with a video model~\cite{yu2025viewcrafter,ren2025gen3c,huang2025voyager}.
Implicit methods encode camera geometry through positional encodings or additional modules.
PE-Field assigns source tokens depth-aware rotary encodings based on their reprojected locations and depths~\cite{bai2025pefield}.
MetaView encodes camera parameters and metric depth in modified rotary embeddings and injects features from a geometry encoder through parallel attention layers added to a frozen pretrained image generator~\cite{cai2026metaview}.

For dynamic NVS, GCD translates the source video to the target viewpoints from relative camera poses, and does not take depth as input~\cite{vanhoorick2024gcd}.
TrajectoryCrafter lifts a monocular video by depth, warps it onto the target trajectory, and completes the result with a diffusion model~\cite{yu2025trajcrafter}.
CogNVS reconstructs the regions observed in the source video, inpaints the unobserved ones, and finetunes the generator at test time~\cite{chen2025cognvs}.
AnyView conditions a video diffusion transformer on Pl\"ucker embeddings of the two trajectories, and omits a depth warp to avoid compounding depth errors and to allow views that overlap little~\cite{vanhoorick2026anyview}.

\subsection{Structured Initial Noise in Diffusion Models}
\label{sec:related-noise}

Information about the target can also enter through the initial sample, since a diffusion model fixes the coarse layout in its first denoising steps.
Warped noise is transported along an optical flow for temporal consistency or motion control, but carries no image content~\cite{chang2025hiwyn,burgert2025gowiththeflow}.
Concurrent works construct geometrically correlated noise with camera geometry or 3D tracks for camera control and multi-view video~\cite{vu2026unicamo,zanjani2026syncvid}; like flow-warped noise, it carries correspondence but no image content.
Other methods build the initial sample from the source: they lower the noise level where the warp explains a pixel~\cite{kong2025worldwarp}, invert the warped video and sample the unobserved regions~\cite{yesiltepe2025inversedvs}, translate the inverted source latent with a learned network~\cite{singh2025ddimnvs}, or learn to map random noise to better initial noise~\cite{zhang2025hqnoise}.
WAVE, which also targets single-image novel view synthesis, mixes the low frequencies of the warp with high-frequency noise~\cite{park2025wave}.
These methods do not explicitly use epipolar geometry to select which frequency orientations retain source phase.

Phase-preserving diffusion instead constructs noise using input phase and random magnitudes, and trains the generator on this structured noise without modifying its architecture.
NeuralRemaster blends input and noise phase using a smooth radial frequency mask~\cite{zeng2025neuralremaster}, and Wavelet Phase Diffusion supports a spatially varying cutoff in a complex wavelet packet transform~\cite{wang2026psipd}.
Both target structure-aligned translation, where the output keeps the viewpoint of the input, so the phase of the input can be trusted where it is kept.
In both methods, the cutoff does not vary independently with frequency orientation.
In contrast, we release the phase of the warped source along the epipolar line where the depth is uncertain, keep it across the line, and do not carry it over into disoccluded regions.
We isolate the effect of directional phase preservation in our generator using the isotropic variant examined in \cref{sec:ablation}.

\section{Preliminaries}
\label{sec:prelim}

Wavelet Phase Diffusion~\cite{wang2026psipd} employs a dual-tree complex wavelet packet transform (\dtcwpt)~\cite{bayram2008dtcwpt}.
Similar to the Fourier transform, the~\dtcwpt{} assigns coefficients to frequency content.
However, a Fourier coefficient covers the whole image with one frequency, while a~\dtcwpt{} coefficient covers a small neighborhood with one frequency band.
The dual-tree structure gives each complex coefficient one orientation, so structures at different angles fall into different coefficients~\cite{selesnick2005dtcwt}.
A packet split divides one such frequency band into narrower bands.
The band of lowest frequency is one real packet $\mathbf{L}$.
Band $l$ is responsible for one frequency interval, centered at $\mathbf{f}_l\in[-1,1]^2$ in Nyquist units, and has complex coefficients $\mathbf{H}_l(\mathbf{q}) = |\mathbf{H}_l(\mathbf{q})| e^{j\angle \mathbf{H}_l(\mathbf{q})}$, each with a magnitude and a phase, indexed by their location $\mathbf{q}$ within the band.

\begin{figure*}[t]
  \centering
  \includegraphics[width=\textwidth]{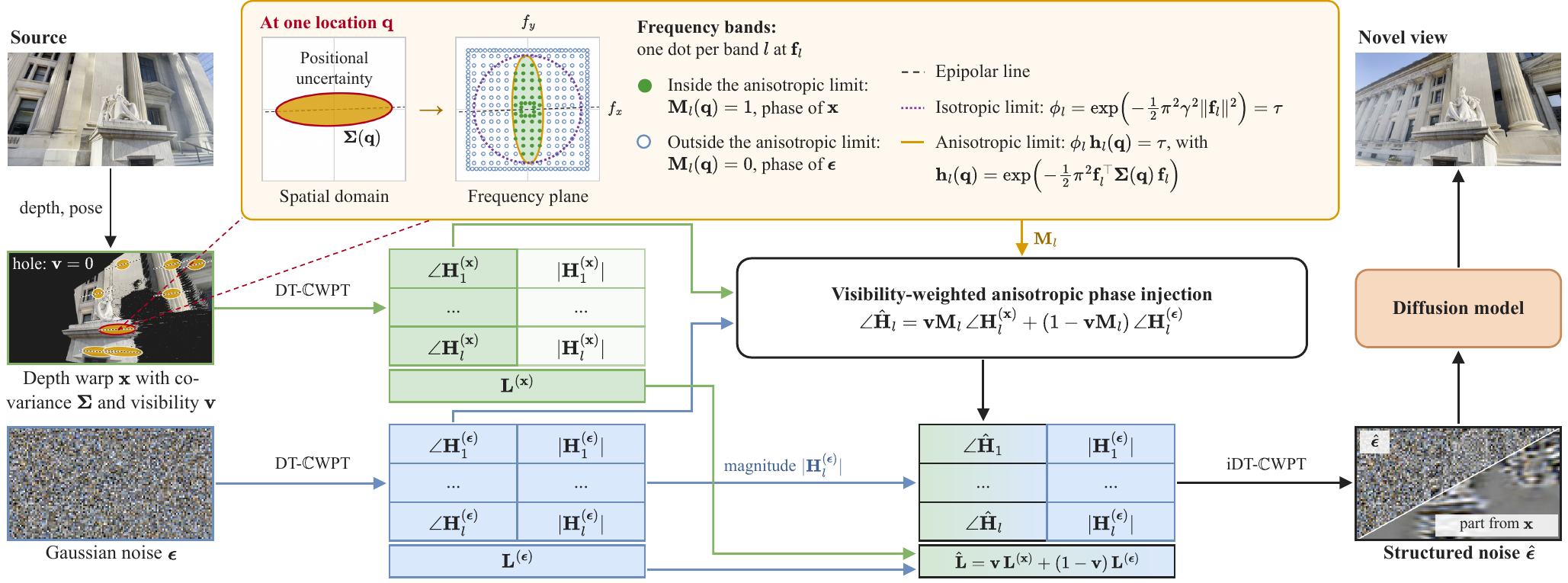}
  \caption{\textbf{Overview of WaveView.}
  The depth warp $\mathbf{x}$ of the source (green) and Gaussian noise $\boldsymbol{\epsilon}$ (blue) are decomposed with the~\dtcwpt{}.
  At a visible location $\mathbf{q}$, the positional uncertainty $\mathbf{\Sigma}(\mathbf{q})$ is stretched along the epipolar line, so bands whose frequency lies along this line take their phase from the noise, and bands keeping the warp's phase (green dots) form an ellipse elongated across it.
  The magnitude of every complex band comes from the noise; holes ($\mathbf{v}=0$) are pure noise.
  Lower right of $\hat{\boldsymbol{\epsilon}}$: the part whose phase comes from the warp.
  }
  \label{fig:method}
\end{figure*}

Wavelet Phase Diffusion decomposes a source $\mathbf{x}$ and a Gaussian noise $\boldsymbol{\epsilon}$, edits the coefficients, and inverts the transform to obtain mixed latent space noise.
The magnitude of $\hat{\mathbf{H}}_l$ is taken from $\boldsymbol{\epsilon}$.
Its phase is taken from $\mathbf{x}$ when $\|\mathbf{f}_l\|$ lies below a cutoff $\mathbf{F}$, and from $\boldsymbol{\epsilon}$ otherwise,
\begin{equation}
\hat{\mathbf{H}}_l = |\mathbf{H}_l^{(\boldsymbol{\epsilon})}| \exp\bigl(j (\mathbf{M}_l \angle \mathbf{H}_l^{(\mathbf{x})} + (1-\mathbf{M}_l)\angle \mathbf{H}_l^{(\boldsymbol{\epsilon})})\bigr),
\end{equation}
where the mask $\mathbf{M}_l(\mathbf{q})\in\{0,1\}$ equals one where $\|\mathbf{f}_l\|\le \mathbf{F}(\mathbf{q})$ and zero otherwise.
The cutoff may change from one location to another thanks to the finite support of the~\dtcwpt{}.
However, at each location, bands in every direction share that cutoff.
In the frequency plane, the retained bands therefore fill a disk of radius $\mathbf{F}(\mathbf{q})$ (dotted circle in \cref{fig:method}).
The inverse~\dtcwpt{} gives the structured noise %
\begin{equation}
\hat{\boldsymbol{\epsilon}} = \text{i}\dtcwpt\bigl(\mathbf{L}^{(\boldsymbol{\epsilon})}, \{\hat{\mathbf{H}}_l\}_l\bigr),
\end{equation}
used as the initial state for the reverse diffusion process.
\section{WaveView}
\label{sec:waveview}

\subsection{Problem Formulation}
\label{sec:problem}

We formulate static NVS from one image and dynamic NVS from a monocular video.
For static NVS, the source and the target are $\mathbf{I}^{\{\mathrm{src},\mathrm{tgt}\}}\in\mathbb{R}^{H\times W\times 3}$, observed and synthesized at cameras $(\mathbf{K}^{\mathrm{src}}, \mathbf{T}^{\mathrm{src}})$ and $(\mathbf{K}^{\mathrm{tgt}}, \mathbf{T}^{\mathrm{tgt}})$, where $\mathbf{K}^{\{\mathrm{src},\mathrm{tgt}\}}\in\mathbb{R}^{3\times 3}$ are intrinsics and $\mathbf{T}^{\{\mathrm{src},\mathrm{tgt}\}}\in\mathbb{R}^{4\times 4}$ are extrinsics.
The relative rigid motion from the source to the target camera is $(\mathbf{R}, \mathbf{t})$, with $\mathbf{R}\in\mathbb{R}^{3\times 3}$ and $\mathbf{t}\in\mathbb{R}^{3}$ given by $\mathbf{T}^{\mathrm{tgt}}(\mathbf{T}^{\mathrm{src}})^{-1}$.
For dynamic NVS from a monocular video of $N$ frames, the target is a second trajectory of the same length, and each quantity gains a leading dimension $N$.
 
\subsection{Positional Uncertainty of the Warp}
\label{sec:cov}

A monocular depth estimator assigns each source pixel $(u^{\mathrm{src}},v^{\mathrm{src}})$ a depth $d^{\mathrm{src}}>0$ along the optical axis.
Lifting the pixel along its ray $\mathbf{r}^{\mathrm{src}}=(\mathbf{K}^{\mathrm{src}})^{-1}(u^{\mathrm{src}},v^{\mathrm{src}},1)^{\top}$ and moving it with $(\mathbf{R},\mathbf{t})$ gives the point in the target camera,
\begin{equation}
\mathbf{X}^{\mathrm{tgt}} = d^{\mathrm{src}}\, \mathbf{R}\, \mathbf{r}^{\mathrm{src}} + \mathbf{t},
\end{equation}
which projects to the target pixel $(u^{\mathrm{tgt}},v^{\mathrm{tgt}})$ with $(u^{\mathrm{tgt}},v^{\mathrm{tgt}},1)^{\top}\propto\mathbf{K}^{\mathrm{tgt}}\mathbf{X}^{\mathrm{tgt}}$.
The estimated depth carries a local error, which we model as zero-mean with the relative standard deviation $\rho\,d^{\mathrm{src}}$; we validate this form on ground-truth depth (see the supplementary material).
To first order, the error moves $\mathbf{X}^{\mathrm{tgt}}$ along the source ray by
\begin{equation}
\label{eqn:sigma_depth}
\boldsymbol{\sigma}_{\mathrm{depth}} = \rho\, d^{\mathrm{src}}\, \mathbf{R}\, \mathbf{r}^{\mathrm{src}}.
\end{equation}
Let $\mathbf{J}=\partial(u^{\mathrm{tgt}},v^{\mathrm{tgt}})/\partial\mathbf{X}^{\mathrm{tgt}}$ be the Jacobian of the target projection, and $\eta$ be the stride of the image encoder.
All covariances below are expressed in latent pixels.
The image of the perturbation, $\mathbf{J}\boldsymbol{\sigma}_{\mathrm{depth}}$, lies on the epipolar line of the pixel (see the supplementary material).
The positional covariance at a target location is the rank-one matrix
\begin{equation}
\label{eqn:Sigma}
\mathbf{\Sigma} = (\mathbf{J}\boldsymbol{\sigma}_{\mathrm{depth}})(\mathbf{J}\boldsymbol{\sigma}_{\mathrm{depth}})^{\top}/\eta^2,
\end{equation}
which has zero variance perpendicular to the epipolar line.
This is the first-order propagation of a 3D covariance through the projection, as in EWA splatting~\cite{zwicker2002ewa}, applied to a covariance supported on a single ray.

\subsection{Anisotropic Phase Injection}
\label{sec:phase}

Let the source image warped onto the target grid (\cref{sec:cov}) be the source signal $\mathbf{x}$ (\cref{sec:prelim}), with visibility $\mathbf{v}(\mathbf{q})\in\{0,1\}$ and covariance $\mathbf{\Sigma}(\mathbf{q})$ at each target location $\mathbf{q}$.
For every location and band, we decide whether the phase of the warp can be trusted, as drawn for one location in the top box of \cref{fig:method}.
Two terms, each with one hyperparameter, enter this decision: $\rho$ releases phase along the epipolar line, and $\gamma$ adds an isotropic floor.

\paragraph{Anisotropic term ($\rho$).}
To first order, for a shift that is small against the support of the packet, shifting the local content by $\boldsymbol{\delta}$ rotates the phase of band $l$ by $\pi\,\mathbf{f}_l\cdot\boldsymbol{\delta}$ and leaves its magnitude unchanged.
For an uncertain shift $\boldsymbol{\delta}\sim\mathcal{N}(\mathbf{0},\mathbf{\Sigma}(\mathbf{q}))$, the coherence is
\begin{equation}
\mathbf{h}_l(\mathbf{q}) = \exp\bigl(-\tfrac12\pi^2 \mathbf{f}_l^{\top} \mathbf{\Sigma}(\mathbf{q})\, \mathbf{f}_l\bigr).
\end{equation}
According to \cref{eqn:sigma_depth,eqn:Sigma}, $\mathbf{\Sigma}(\mathbf{q})$ scales with $\rho^2$ and is stretched along the epipolar line, $\rho$ lowers the coherence only of bands whose frequency lies along that line and leaves bands across it untouched.

\paragraph{Isotropic floor ($\gamma$).}
The warp itself may not be exact in any direction due to camera parameter inaccuracy, so we add the coherence of an isotropic shift of scale $\gamma$,
\begin{equation}
\phi_l=\exp\bigl(-\tfrac12\pi^2 \gamma^2\, \|\mathbf{f}_l\|^2\bigr),
\end{equation}
which lowers high frequencies in every direction, like the cutoff $\mathbf{F}$ of Wavelet Phase Diffusion.

\paragraph{Injection.}
The phase of the warp is kept where both terms together exceed a threshold $\tau$,
\begin{equation}
\mathbf{M}_l(\mathbf{q}) = \begin{cases}1 & \text{if } \phi_l\,\mathbf{h}_l(\mathbf{q})\ge\tau,\\ 0 & \text{otherwise.}\end{cases}
\end{equation}
In the frequency plane, the retained bands (green dots) thus fill an ellipse inside the circle $\phi_l=\tau$ (dotted): $\gamma$ alone sets its major axis, across the epipolar line, where the ellipse touches the circle, and $\rho$ further shortens its minor axis, along the epipolar line (orange).

\paragraph{Visibility ($\mathbf{v}$).}
Only visible locations take the phase of the warp, so the mask of \cref{sec:prelim} becomes $\mathbf{v}\mathbf{M}_l$,
\begin{equation}
\angle\hat{\mathbf{H}}_l = \mathbf{v} \mathbf{M}_l\, \angle \mathbf{H}_l^{(\mathbf{x})} + (1-\mathbf{v} \mathbf{M}_l)\, \angle \mathbf{H}_l^{(\boldsymbol{\epsilon})},
\end{equation}
and the magnitude of every band is that of the noise.
The lowest packet $\mathbf{L}$ is real-valued, so we mix the whole packet from the warp and noise according to visibility,
\begin{equation}
\hat{\mathbf{L}} = \mathbf{v}\,\mathbf{L}^{(\mathbf{x})} + (1-\mathbf{v})\,\mathbf{L}^{(\boldsymbol{\epsilon})},
\end{equation}
where $\mathbf{L}^{(\mathbf{x})}$ is first standardized to the mean and standard deviation of $\mathbf{L}^{(\boldsymbol{\epsilon})}$, so that $\hat{\mathbf{L}}$ carries the coarse layout of the warp with the statistics of the noise.
Further details are given in the supplementary material.

\begin{table*}[t]
  \centering
  \caption{\textbf{Quantitative results on DL3DV}. Covisible region: target pixels covisible with the source. Best in bold, second underlined.}
  \label{tab:main}
  \scriptsize
  \setlength{\tabcolsep}{1.6pt}
  \begin{tabular}{@{}l*{6}{ccc}@{}}
    \toprule
    & \multicolumn{6}{c}{Easy (overlap $>$80\%)} & \multicolumn{6}{c}{Medium (50--80\%)} & \multicolumn{6}{c}{Hard (30--50\%)} \\
    \cmidrule(lr){2-7}\cmidrule(lr){8-13}\cmidrule(lr){14-19}
    & \multicolumn{3}{c}{Whole image} & \multicolumn{3}{c}{Covisible region} & \multicolumn{3}{c}{Whole image} & \multicolumn{3}{c}{Covisible region} & \multicolumn{3}{c}{Whole image} & \multicolumn{3}{c}{Covisible region} \\
    \cmidrule(lr){2-4}\cmidrule(lr){5-7}\cmidrule(lr){8-10}\cmidrule(lr){11-13}\cmidrule(lr){14-16}\cmidrule(lr){17-19}
    Method & PSNR$\uparrow$ & SSIM$\uparrow$ & LPIPS$\downarrow$ & PSNR$\uparrow$ & SSIM$\uparrow$ & LPIPS$\downarrow$ & PSNR$\uparrow$ & SSIM$\uparrow$ & LPIPS$\downarrow$ & PSNR$\uparrow$ & SSIM$\uparrow$ & LPIPS$\downarrow$ & PSNR$\uparrow$ & SSIM$\uparrow$ & LPIPS$\downarrow$ & PSNR$\uparrow$ & SSIM$\uparrow$ & LPIPS$\downarrow$ \\
    \midrule
    Voyager~\cite{huang2025voyager} & 17.91 & 0.514 & 0.425 & 18.60 & 0.502 & 0.412 & 14.16 & 0.400 & 0.521 & 15.57 & 0.409 & 0.483 & 12.08 & 0.348 & 0.596 & 13.33 & 0.350 & 0.556 \\
    PE-Field~\cite{bai2025pefield} & 15.26 & 0.384 & 0.456 & 15.26 & 0.350 & 0.456 & 13.40 & 0.344 & 0.547 & 13.87 & 0.323 & 0.532 & 12.15 & 0.318 & 0.619 & 12.55 & 0.299 & 0.609 \\
    GEN3C~\cite{ren2025gen3c} & 18.90 & \underline{0.603} & 0.301 & 19.64 & \underline{0.603} & 0.279 & 14.45 & 0.414 & 0.452 & 15.68 & 0.429 & 0.409 & 12.61 & 0.324 & 0.546 & 13.60 & 0.325 & 0.508 \\
    ViewCrafter~\cite{yu2025viewcrafter} & 18.89 & 0.593 & 0.324 & 19.66 & 0.585 & 0.305 & 14.66 & \underline{0.431} & 0.482 & 15.88 & \underline{0.439} & 0.441 & 12.88 & \underline{0.367} & 0.578 & 13.95 & \underline{0.366} & 0.539 \\
    MetaView~\cite{cai2026metaview} & \underline{19.09} & 0.590 & \underline{0.282} & \underline{19.72} & 0.583 & \underline{0.262} & \underline{14.94} & 0.421 & \underline{0.441} & \underline{15.99} & 0.423 & \underline{0.400} & \underline{13.12} & 0.353 & \underline{0.540} & \underline{13.98} & 0.349 & \underline{0.501} \\
    WaveView (Ours) & \textbf{20.24} & \textbf{0.654} & \textbf{0.266} & \textbf{21.08} & \textbf{0.656} & \textbf{0.241} & \textbf{15.36} & \textbf{0.458} & \textbf{0.431} & \textbf{16.69} & \textbf{0.478} & \textbf{0.383} & \textbf{13.71} & \textbf{0.386} & \textbf{0.523} & \textbf{14.82} & \textbf{0.401} & \textbf{0.471} \\
    \bottomrule
  \end{tabular}
\end{table*}

\begin{table*}[!t]
  \centering
  \caption{\textbf{Dynamic novel view synthesis on AnyViewBench}~\cite{vanhoorick2026anyview}. Baseline numbers from~\cite{vanhoorick2026anyview}. Best in bold, second underlined. $^\ast$ absent from our training data (see~\cref{sec:setup}).}
  \label{tab:anyview}
  \scriptsize
  \setlength{\tabcolsep}{1.4pt}
  \begin{tabular}{@{}l*{6}{rrr}@{}}
    \toprule
    & \multicolumn{3}{c}{GCD~\cite{vanhoorick2024gcd}} & \multicolumn{3}{c}{GEN3C~\cite{ren2025gen3c}} & \multicolumn{3}{c}{CogNVS~\cite{chen2025cognvs}} & \multicolumn{3}{c}{TrajCrafter~\cite{yu2025trajcrafter}} & \multicolumn{3}{c}{AnyView~\cite{vanhoorick2026anyview}} & \multicolumn{3}{c}{WaveView (Ours)} \\
    \cmidrule(lr){2-4}\cmidrule(lr){5-7}\cmidrule(lr){8-10}\cmidrule(lr){11-13}\cmidrule(lr){14-16}\cmidrule(lr){17-19}
    Dataset & PSNR$\uparrow$ & SSIM$\uparrow$ & LPIPS$\downarrow$ & PSNR$\uparrow$ & SSIM$\uparrow$ & LPIPS$\downarrow$ & PSNR$\uparrow$ & SSIM$\uparrow$ & LPIPS$\downarrow$ & PSNR$\uparrow$ & SSIM$\uparrow$ & LPIPS$\downarrow$ & PSNR$\uparrow$ & SSIM$\uparrow$ & LPIPS$\downarrow$ & PSNR$\uparrow$ & SSIM$\uparrow$ & LPIPS$\downarrow$ \\
    \midrule
    \multicolumn{19}{@{}l}{\textit{AnyViewBench: in-distribution}} \\
    DROID~\cite{khazatsky2024droid} & 10.18 & 0.255 & 0.688 & 9.62 & 0.211 & 0.666 & 9.44 & 0.281 & 0.634 & 10.45 & 0.257 & 0.620 & \underline{14.42} & \underline{0.443} & \underline{0.473} & \textbf{14.92} & \textbf{0.460} & \textbf{0.453} \\
    Ego-Exo4D~\cite{grauman2024egoexo4d}$^\ast$ & 12.10 & 0.255 & 0.670 & 11.69 & 0.222 & 0.643 & 10.80 & 0.241 & 0.670 & 11.33 & 0.195 & 0.642 & \textbf{17.46} & \textbf{0.493} & \textbf{0.410} & \underline{16.02} & \underline{0.429} & \underline{0.456} \\
    LBM~\cite{vanhoorick2026anyview}$^\ast$ & 12.59 & 0.398 & 0.694 & 13.48 & 0.449 & 0.581 & 13.30 & 0.453 & 0.548 & 13.68 & 0.447 & 0.537 & \underline{17.63} & \underline{0.619} & \underline{0.386} & \textbf{18.17} & \textbf{0.644} & \textbf{0.371} \\
    Kubric-4D~\cite{vanhoorick2024gcd,greff2022kubric} & 17.57 & 0.477 & 0.512 & 13.59 & 0.341 & 0.599 & 12.55 & 0.329 & 0.601 & 14.14 & 0.294 & 0.592 & \textbf{19.91} & \textbf{0.541} & \textbf{0.427} & \underline{19.45} & \underline{0.538} & \underline{0.443} \\
    Kubric-5D~\cite{vanhoorick2026anyview} & 13.83 & 0.391 & 0.644 & 13.10 & 0.327 & 0.627 & 12.18 & 0.318 & 0.643 & 13.30 & 0.287 & 0.625 & \underline{17.04} & \underline{0.460} & \underline{0.488} & \textbf{17.27} & \textbf{0.478} & \textbf{0.478} \\
    Lyft~\cite{houston2020lyft}$^\ast$ & 8.72 & 0.335 & 0.697 & 8.43 & 0.286 & 0.634 & 8.34 & 0.319 & 0.628 & 8.73 & 0.251 & 0.621 & \textbf{15.29} & \textbf{0.560} & \textbf{0.389} & \underline{14.85} & \underline{0.542} & \underline{0.397} \\
    ParDom-4D~\cite{vanhoorick2024gcd} & 22.67 & \underline{0.656} & \underline{0.457} & 16.64 & 0.478 & 0.590 & 18.36 & 0.499 & 0.564 & 18.23 & 0.475 & 0.586 & \underline{22.85} & 0.641 & 0.469 & \textbf{23.30} & \textbf{0.683} & \textbf{0.438} \\
    Waymo~\cite{sun2020waymo} & 12.93 & \underline{0.393} & 0.647 & 12.98 & 0.350 & 0.600 & 13.27 & 0.377 & 0.594 & 12.66 & 0.312 & 0.593 & \underline{16.31} & \textbf{0.470} & \textbf{0.485} & \textbf{16.53} & \textbf{0.470} & \underline{0.488} \\
    \midrule
    Average & 13.95 & 0.400 & 0.623 & 12.44 & 0.333 & 0.617 & 12.28 & 0.352 & 0.610 & 12.82 & 0.315 & 0.602 & \textbf{17.61} & \underline{0.528} & \underline{0.441} & \underline{17.56} & \textbf{0.530} & \textbf{0.440} \\
    \midrule
    \multicolumn{19}{@{}l}{\textit{AnyViewBench: zero-shot}} \\
    Argoverse~\cite{wilson2021argoverse2} & 11.45 & \underline{0.403} & 0.682 & 10.52 & 0.319 & 0.680 & 10.76 & 0.360 & \underline{0.610} & 10.67 & 0.325 & 0.621 & \underline{11.87} & 0.388 & 0.645 & \textbf{13.44} & \textbf{0.474} & \textbf{0.580} \\
    AssemblyHands~\cite{ohkawa2023assemblyhands} & 9.77 & 0.262 & 0.759 & 9.86 & 0.237 & 0.732 & 9.93 & \underline{0.281} & \underline{0.701} & \textbf{11.45} & 0.248 & \underline{0.701} & 10.39 & 0.275 & 0.711 & \underline{10.93} & \textbf{0.310} & \textbf{0.696} \\
    DDAD~\cite{guizilini2020ddad} & 9.81 & 0.278 & 0.660 & 9.35 & 0.259 & 0.600 & \underline{10.81} & \textbf{0.355} & 0.572 & 10.73 & 0.300 & \underline{0.558} & 10.63 & 0.314 & 0.568 & \textbf{11.22} & \underline{0.346} & \textbf{0.529} \\
    DROID~\cite{khazatsky2024droid} & 11.81 & 0.315 & 0.690 & 10.56 & 0.276 & 0.674 & 10.48 & 0.358 & 0.632 & 11.37 & 0.339 & 0.614 & \underline{11.98} & \underline{0.399} & \underline{0.603} & \textbf{12.83} & \textbf{0.436} & \textbf{0.562} \\
    Ego-Exo4D~\cite{grauman2024egoexo4d} & 11.98 & 0.239 & 0.668 & 11.40 & 0.193 & 0.651 & 10.52 & 0.227 & 0.683 & 11.27 & 0.180 & 0.647 & \underline{13.02} & \underline{0.281} & \underline{0.578} & \textbf{13.31} & \textbf{0.293} & \textbf{0.570} \\
    \midrule
    Average & 10.96 & 0.299 & 0.692 & 10.34 & 0.257 & 0.667 & 10.50 & 0.316 & 0.640 & 11.10 & 0.279 & 0.628 & \underline{11.58} & \underline{0.331} & \underline{0.621} & \textbf{12.35} & \textbf{0.372} & \textbf{0.587} \\
    \bottomrule
  \end{tabular}
\end{table*}

\newlength{\qualw}
\newcommand{\qualpanel}[1]{\includegraphics[width=\qualw]{figures/qual/#1}}
\newcommand{\quallabel}[1]{\raisebox{0.275\qualw}{\rotatebox[origin=c]{90}{\scriptsize #1}}\hspace{3pt}}
\begin{figure*}[t]
  \centering
  \setlength{\qualw}{0.175\textwidth}%
  {\setlength{\tabcolsep}{0pt}\renewcommand{\arraystretch}{0}%
  \begin{tabular}{cccccc}
    \quallabel{Source} & \qualpanel{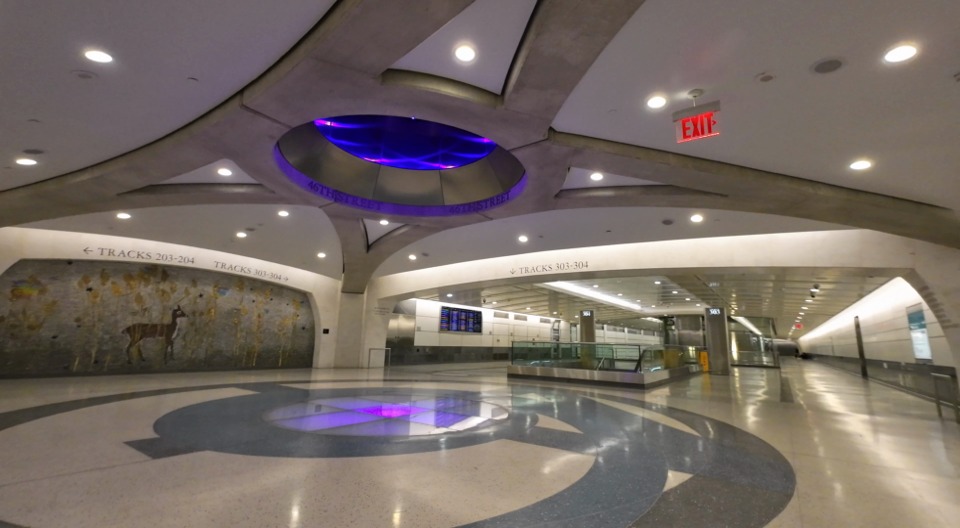} & \qualpanel{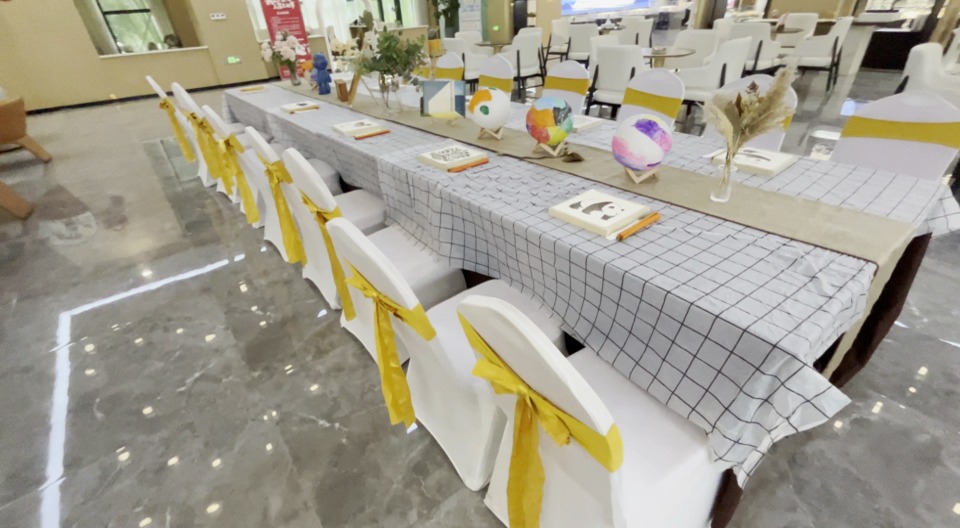} & \qualpanel{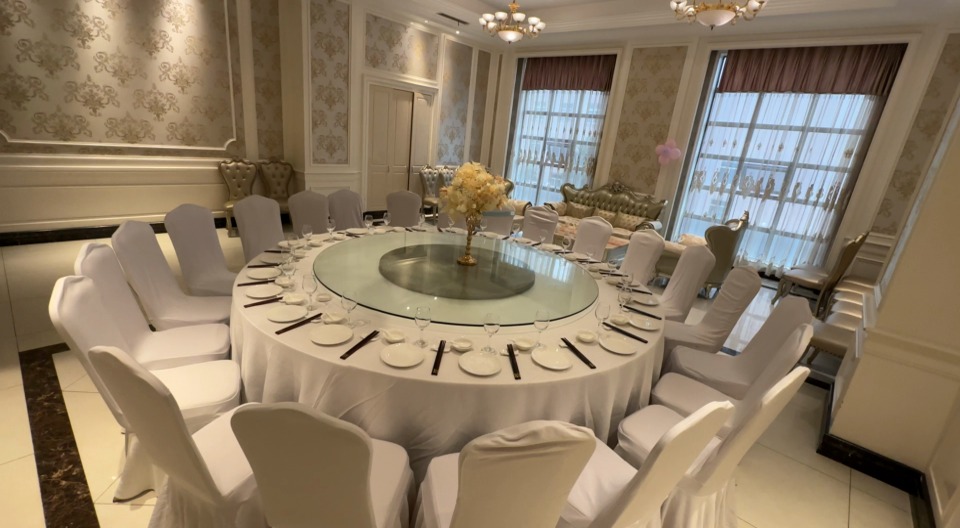} & \qualpanel{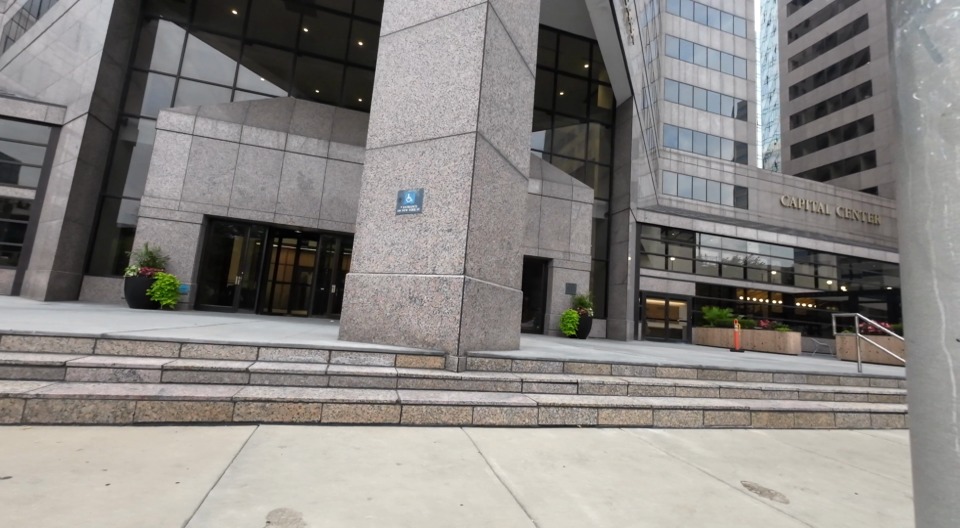} & \qualpanel{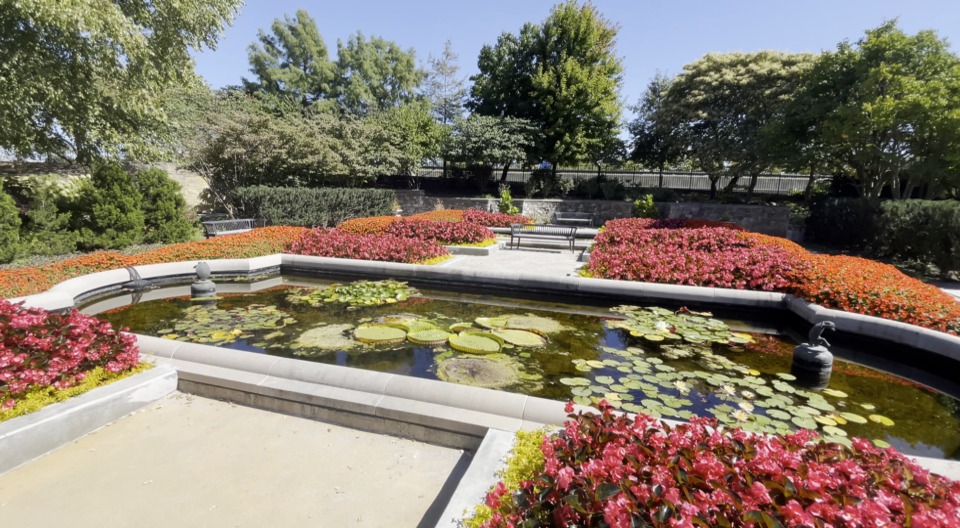} \\
    \quallabel{Warp} & \qualpanel{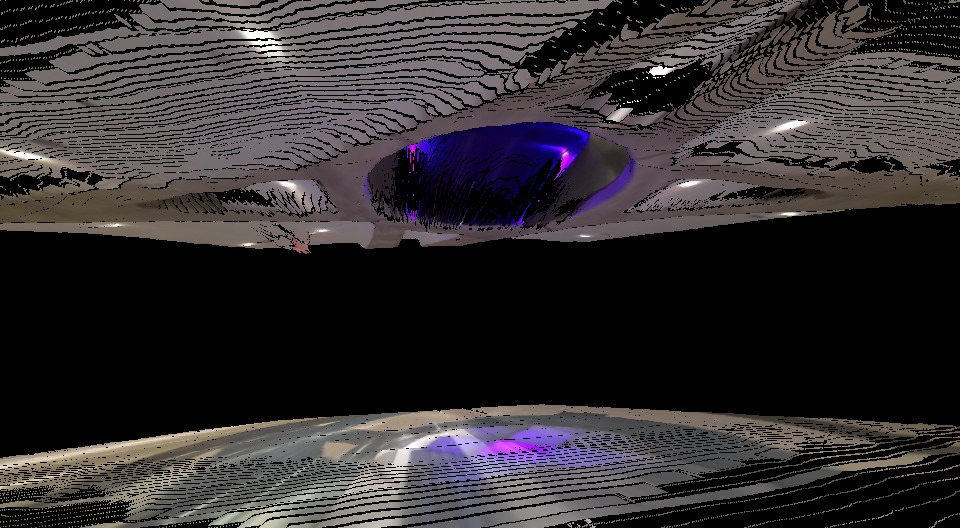} & \qualpanel{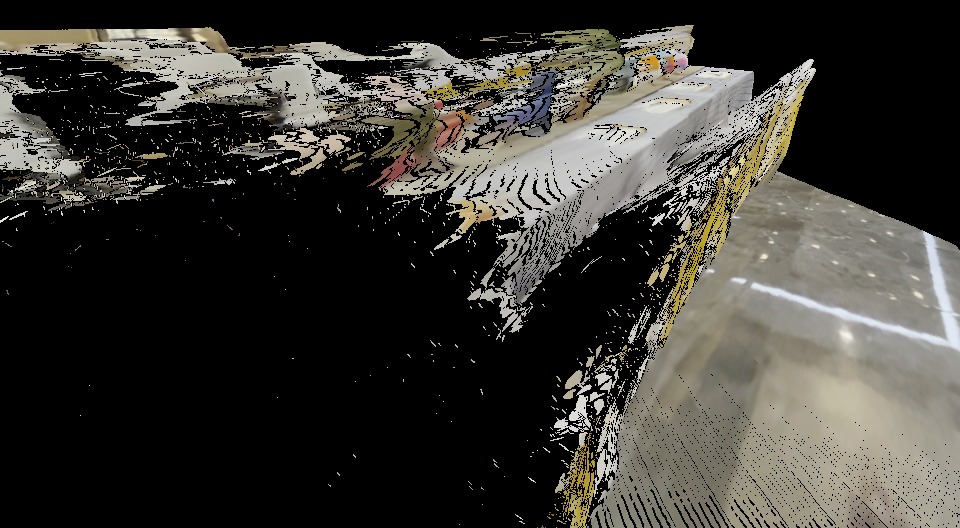} & \qualpanel{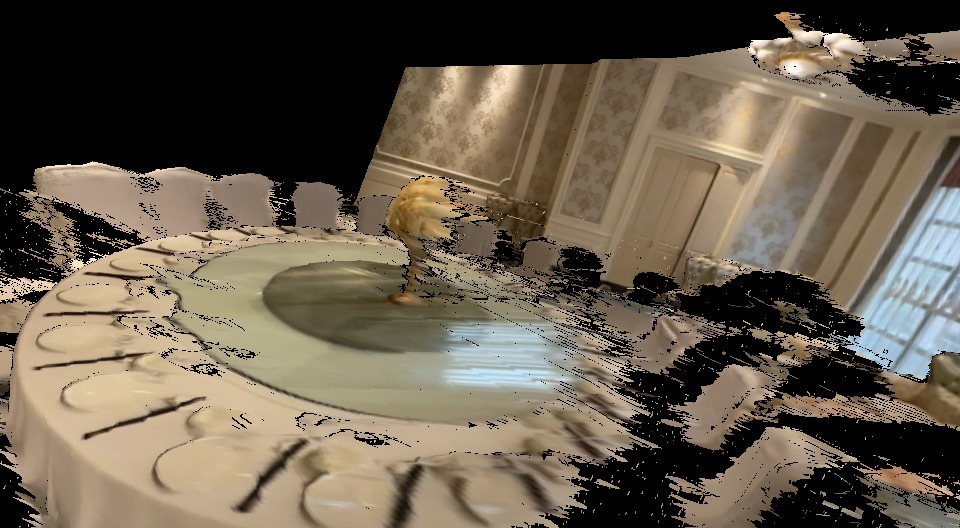} & \qualpanel{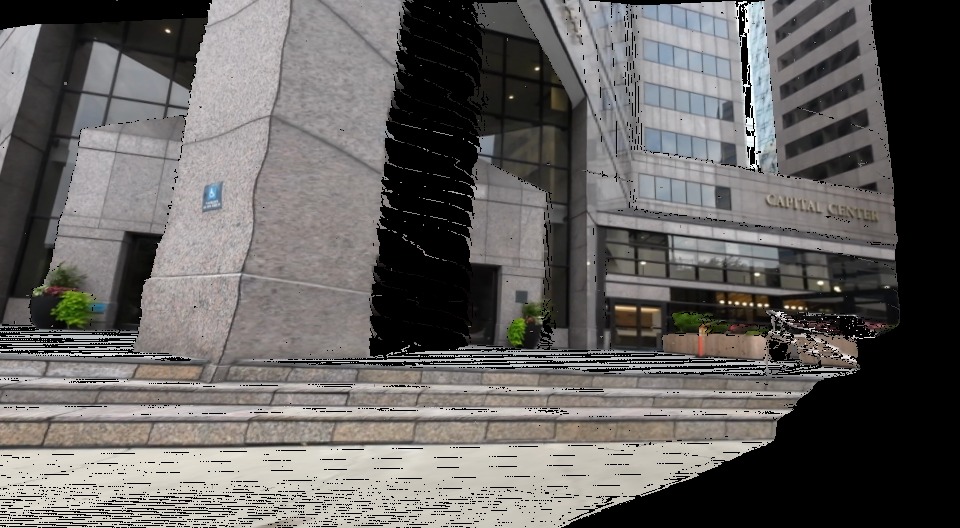} & \qualpanel{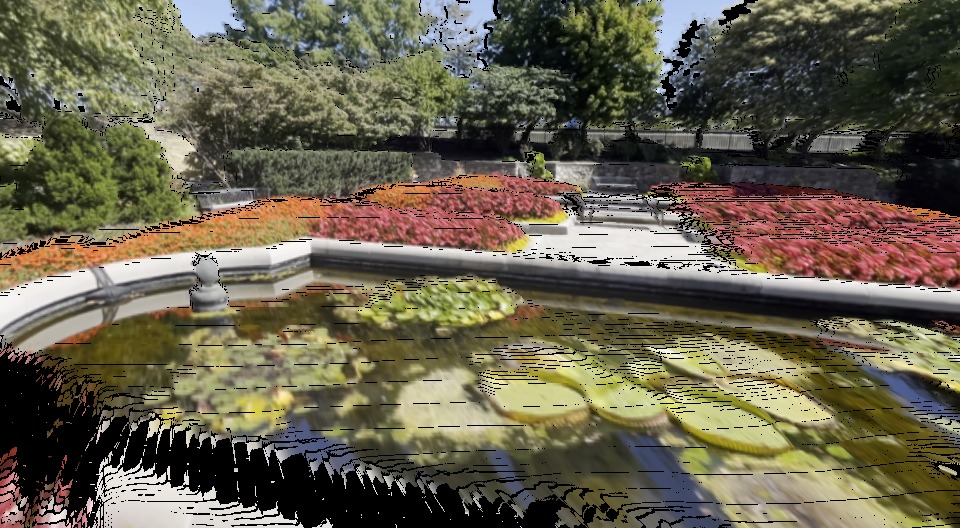} \\[2pt]
    \quallabel{PE-Field} & \qualpanel{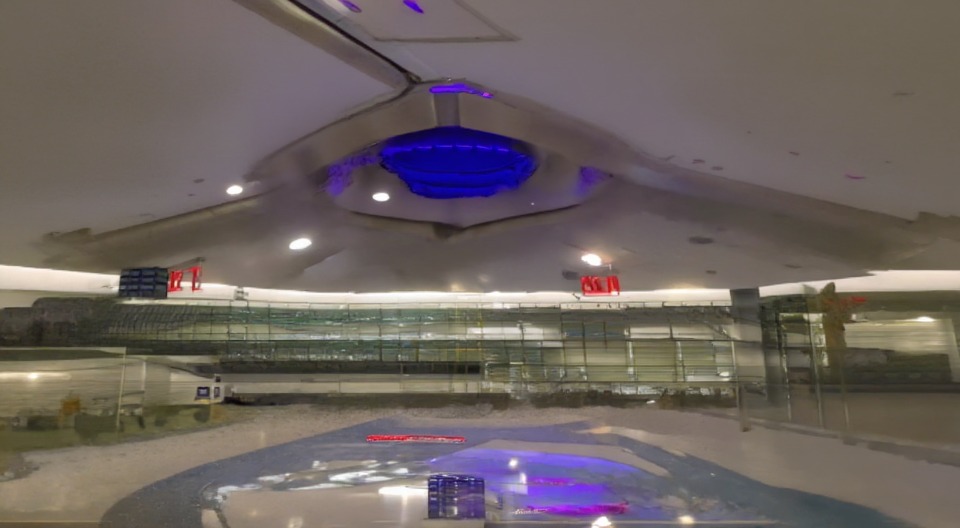} & \qualpanel{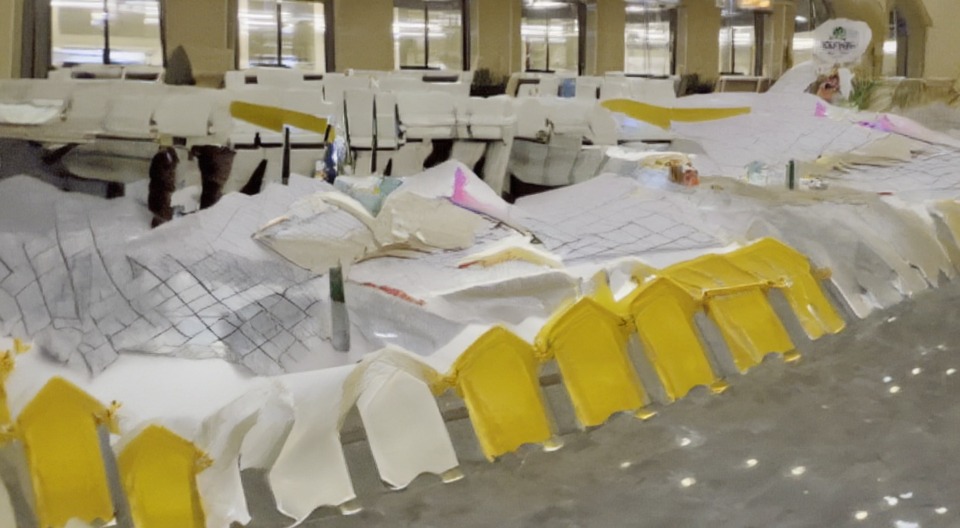} & \qualpanel{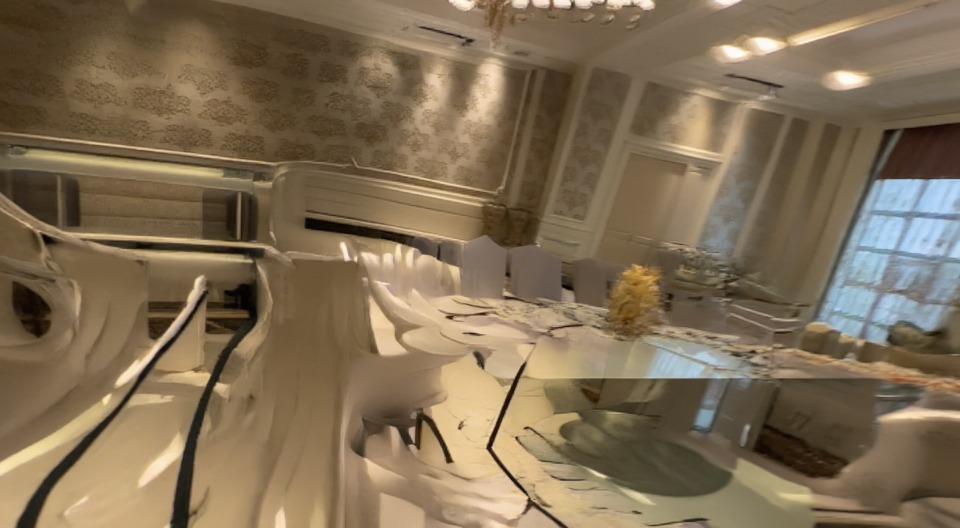} & \qualpanel{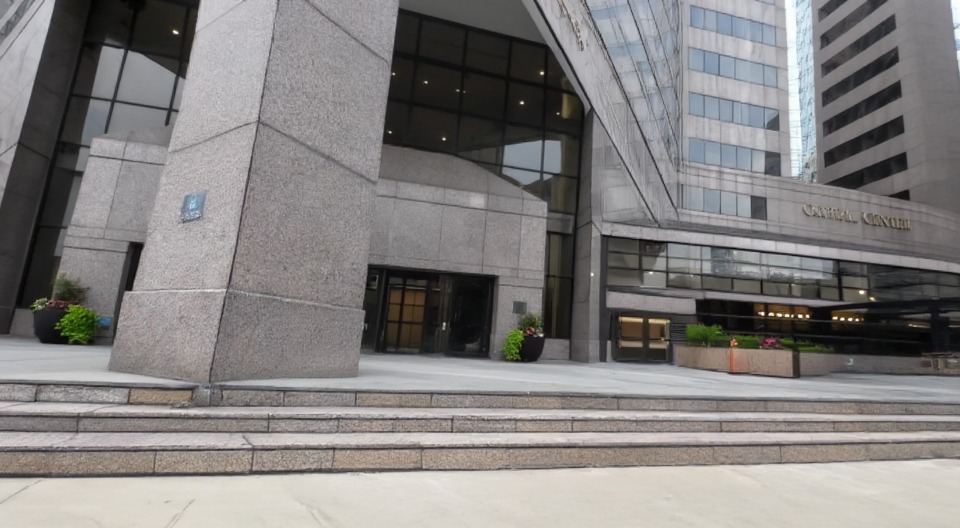} & \qualpanel{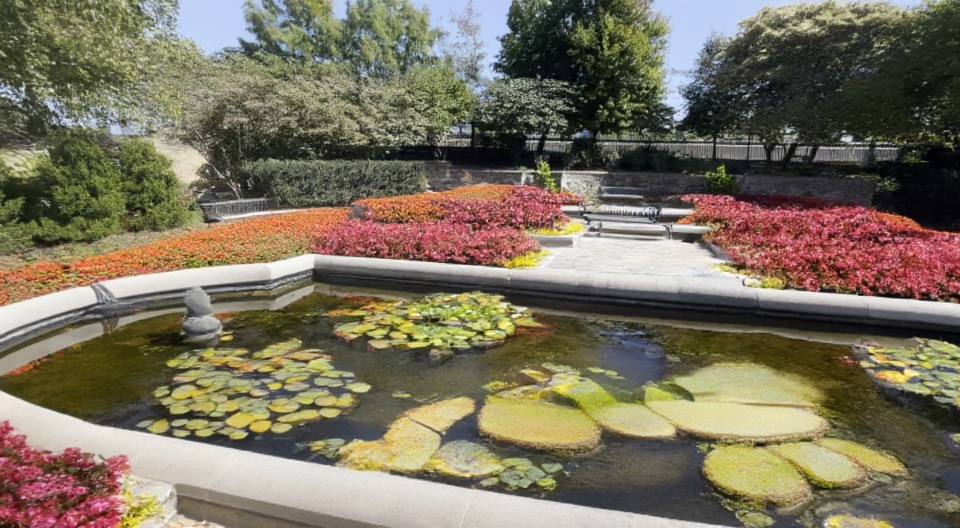} \\
    \quallabel{GEN3C} & \qualpanel{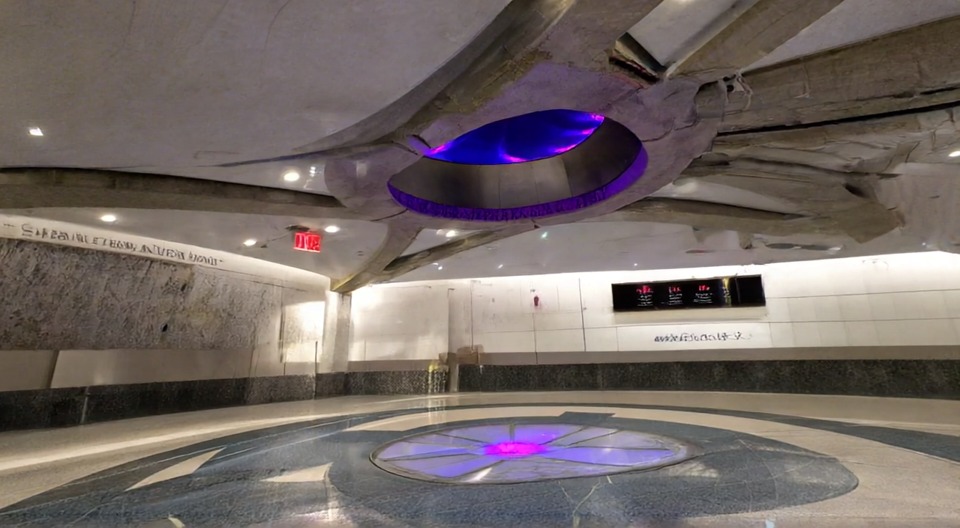} & \qualpanel{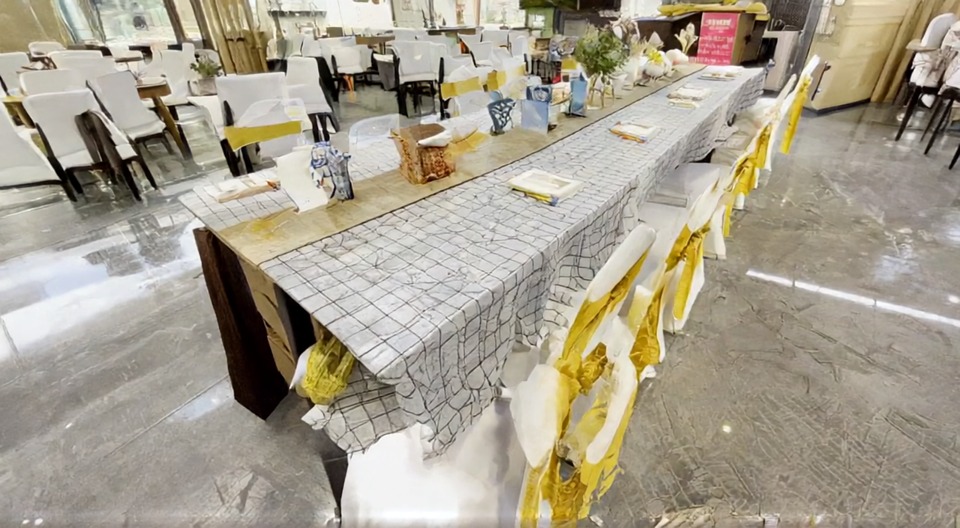} & \qualpanel{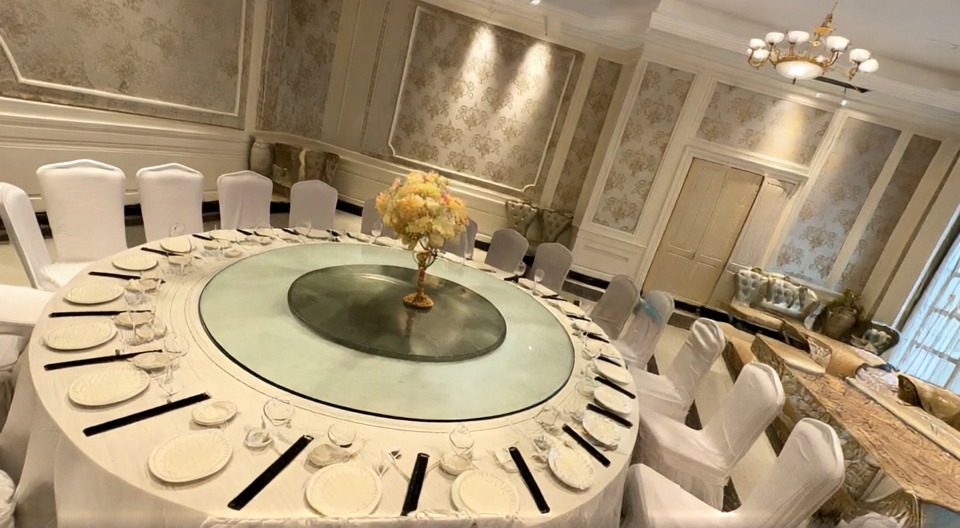} & \qualpanel{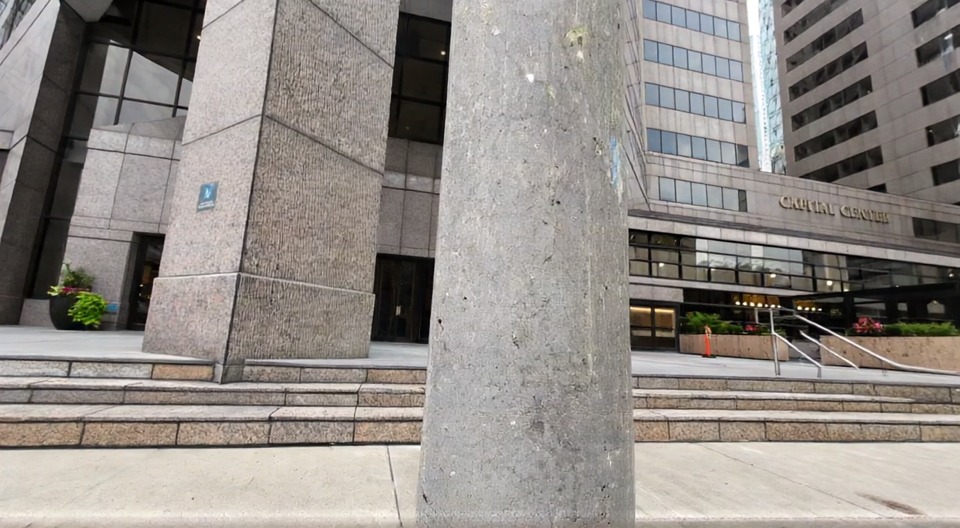} & \qualpanel{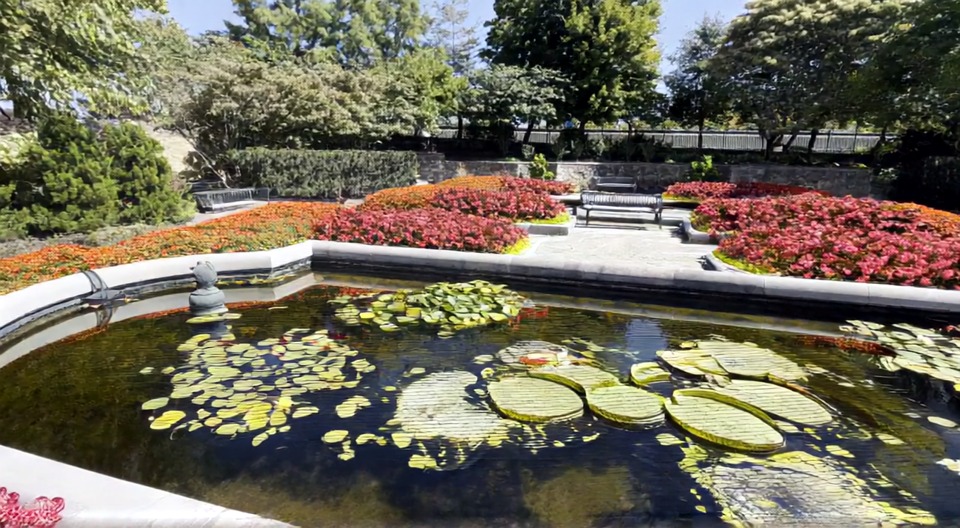} \\
    \quallabel{ViewCrafter} & \qualpanel{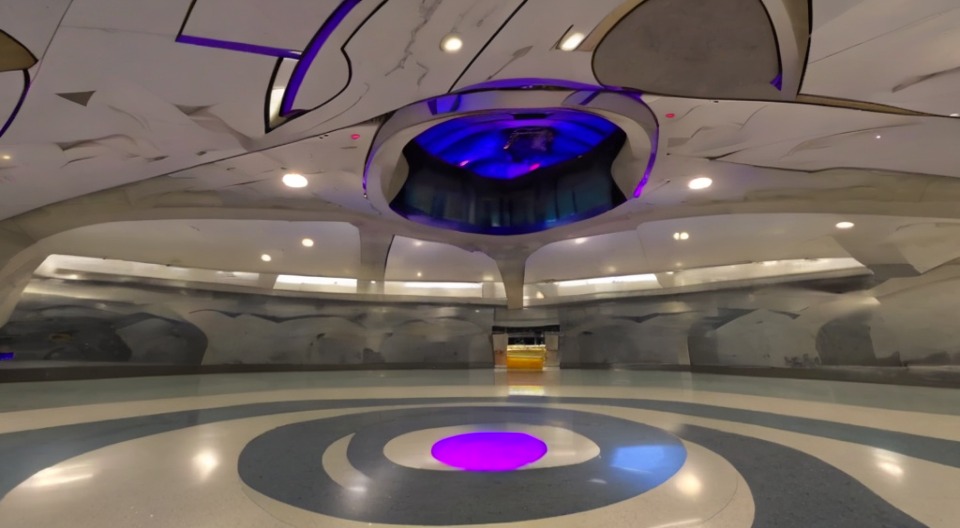} & \qualpanel{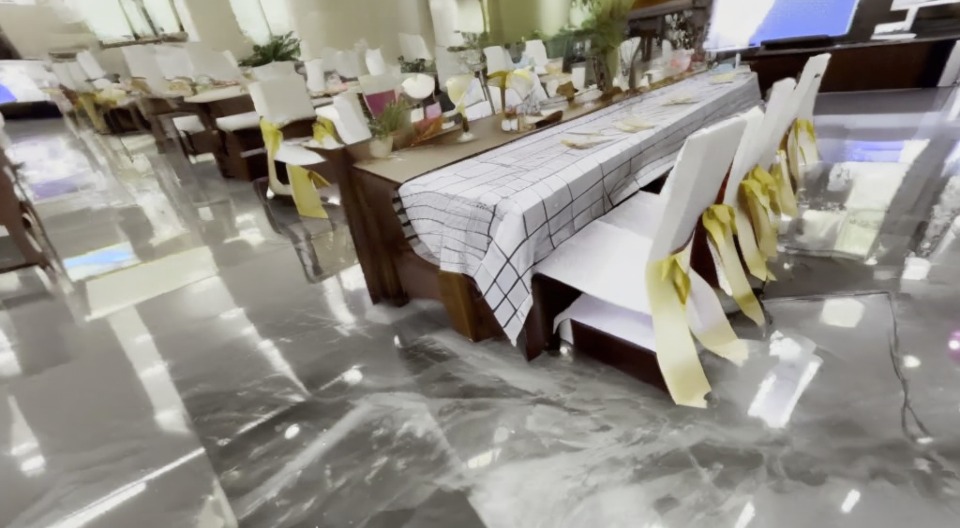} & \qualpanel{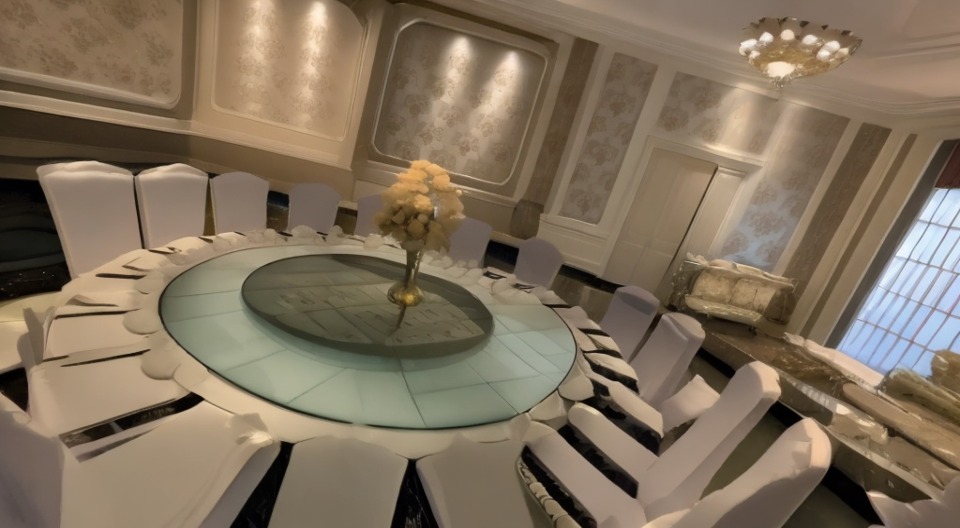} & \qualpanel{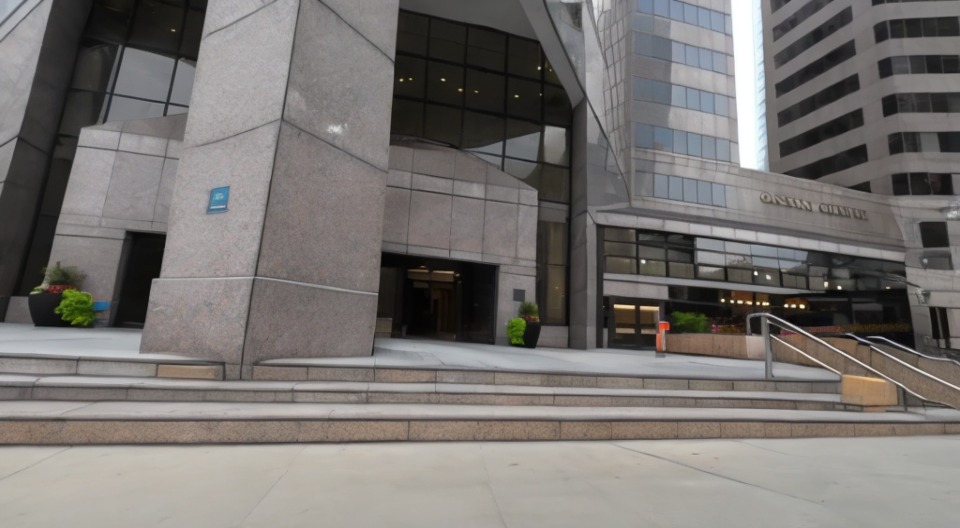} & \qualpanel{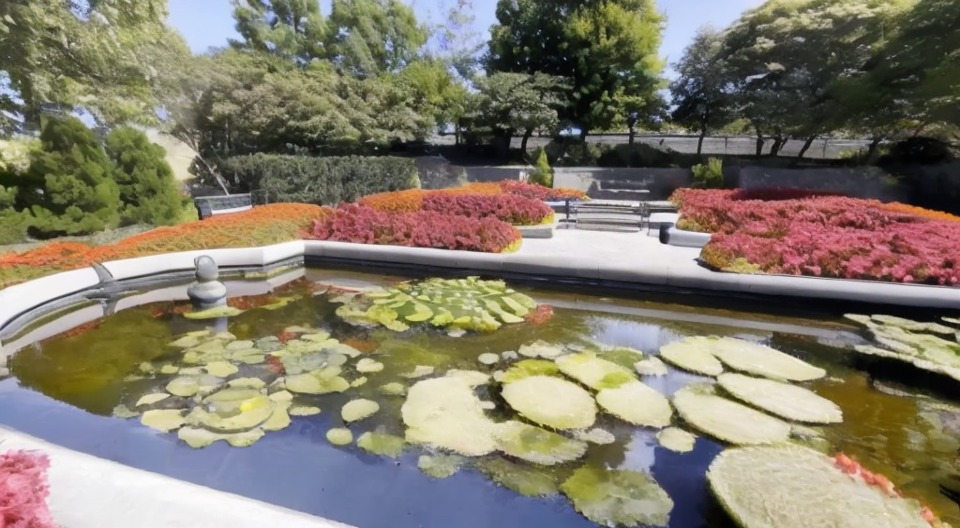} \\
    \quallabel{MetaView} & \qualpanel{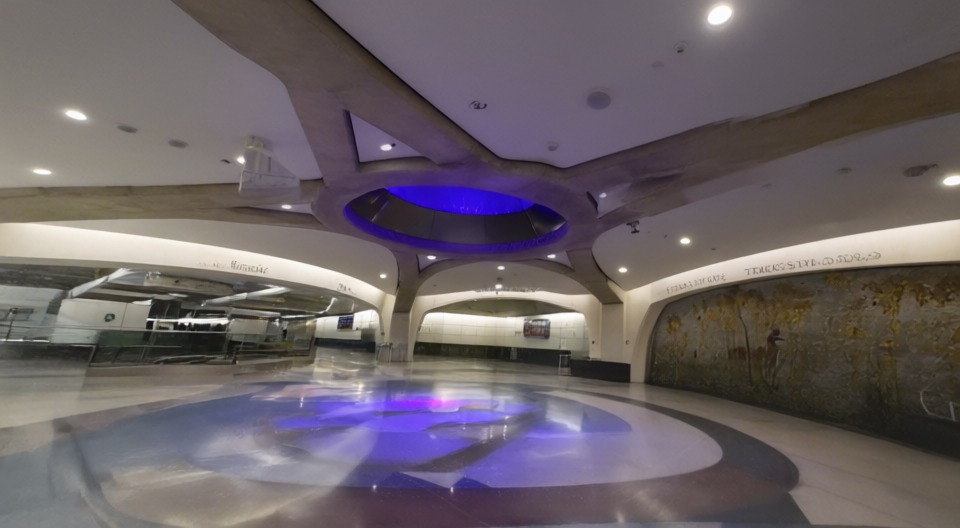} & \qualpanel{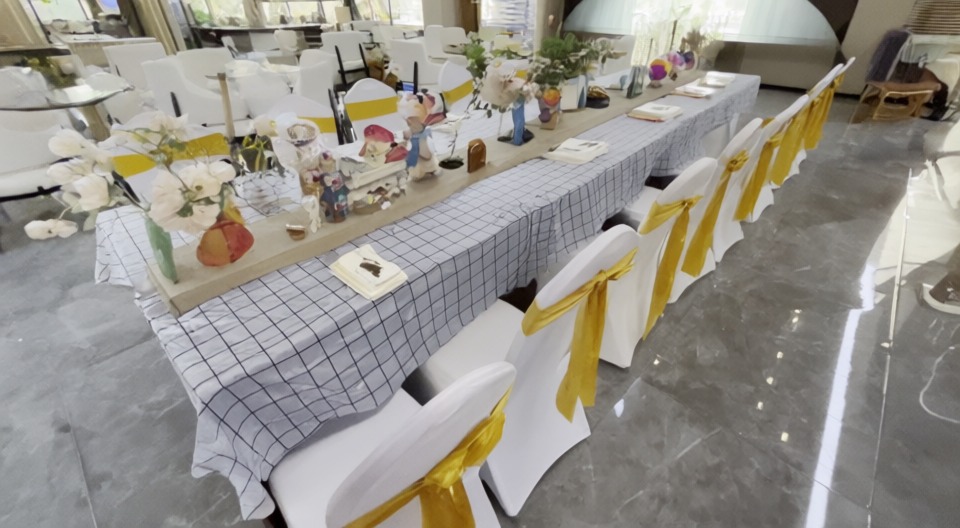} & \qualpanel{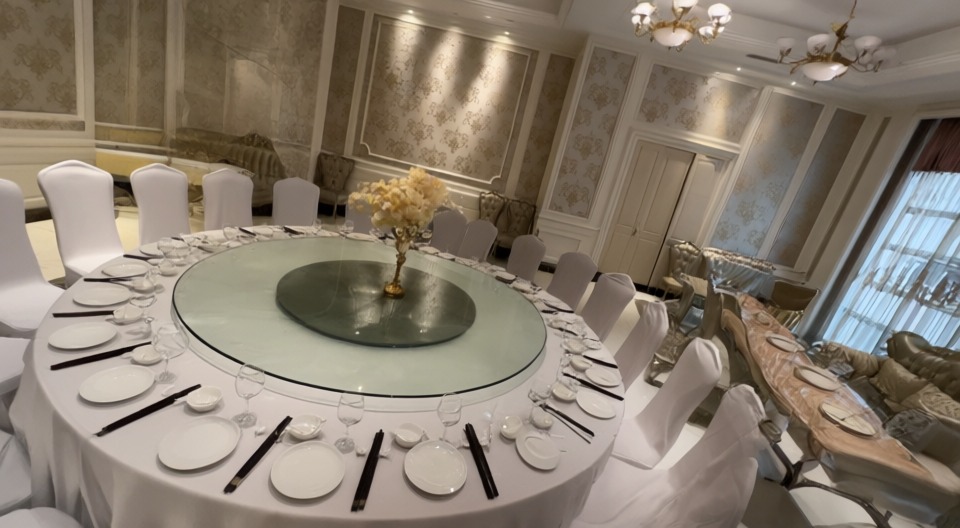} & \qualpanel{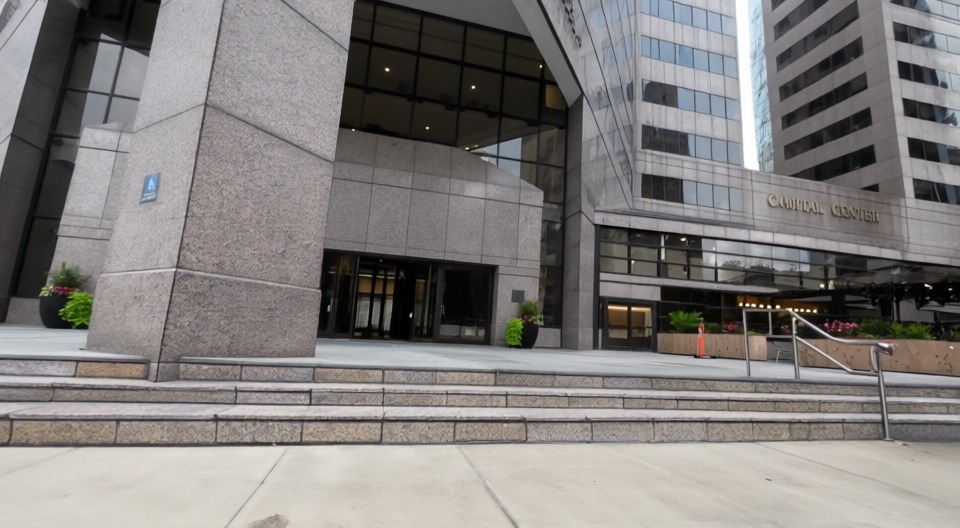} & \qualpanel{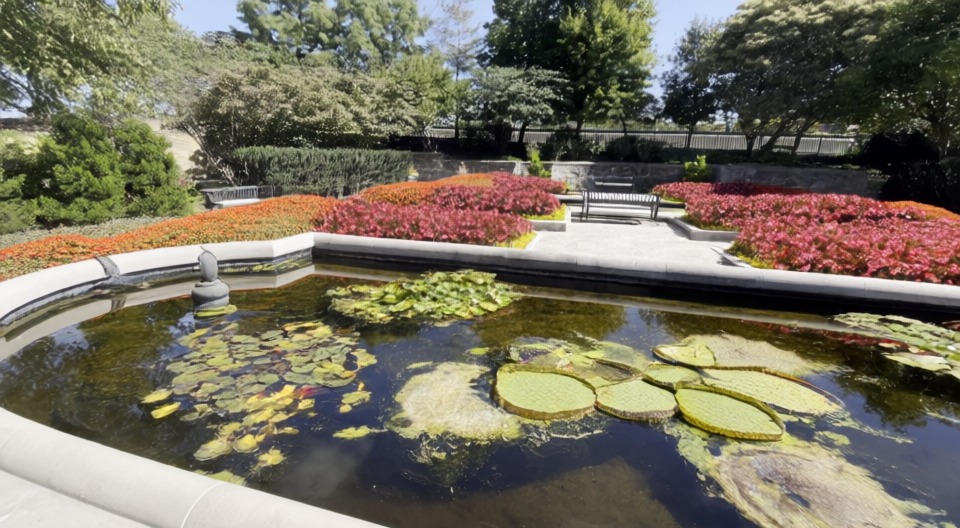} \\
    \quallabel{Ours} & \qualpanel{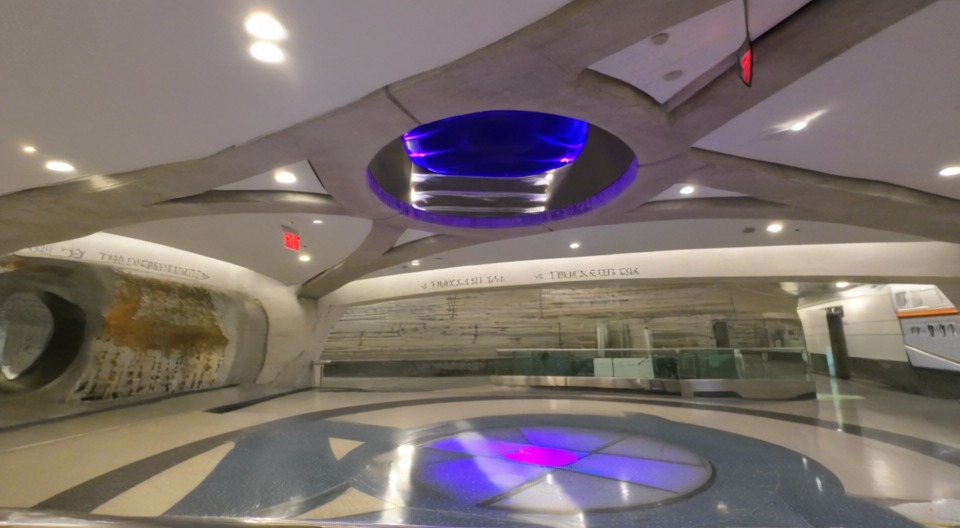} & \qualpanel{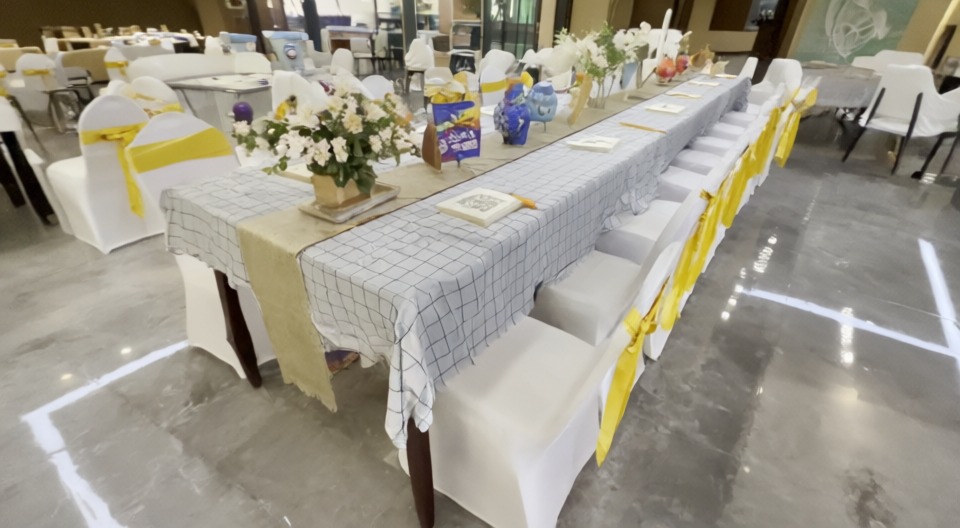} & \qualpanel{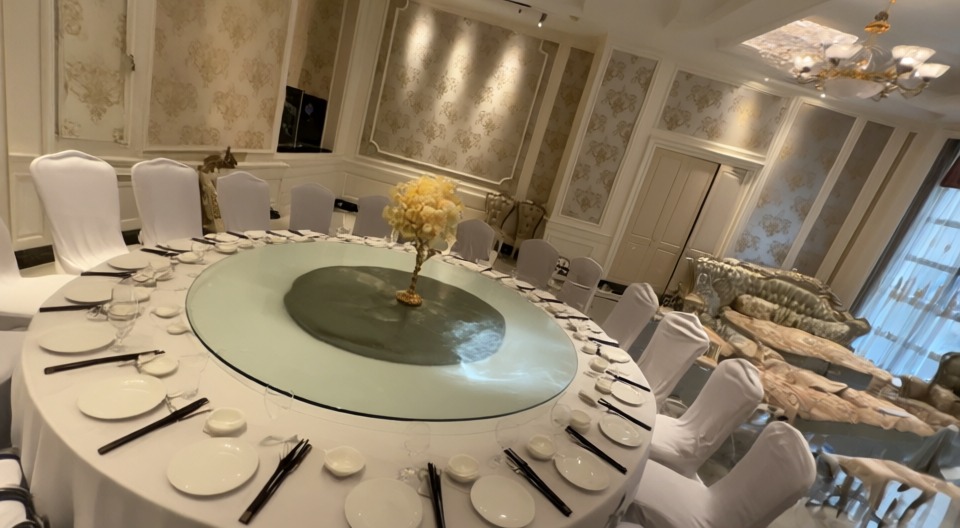} & \qualpanel{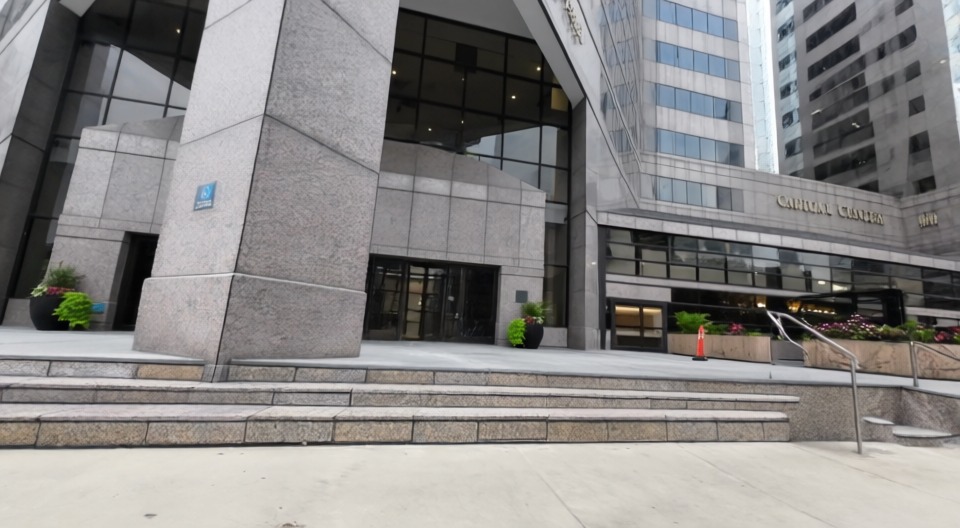} & \qualpanel{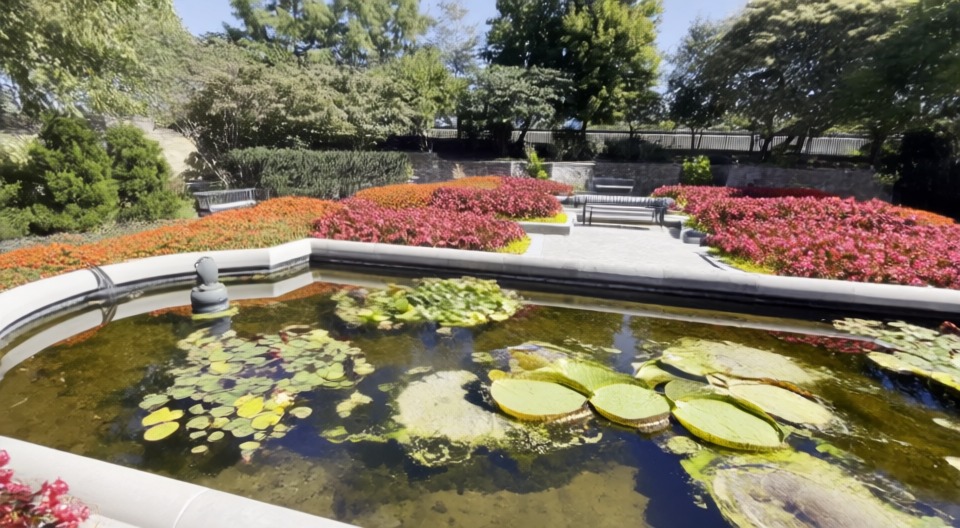} \\[2pt]
    \quallabel{Ground truth} & \qualpanel{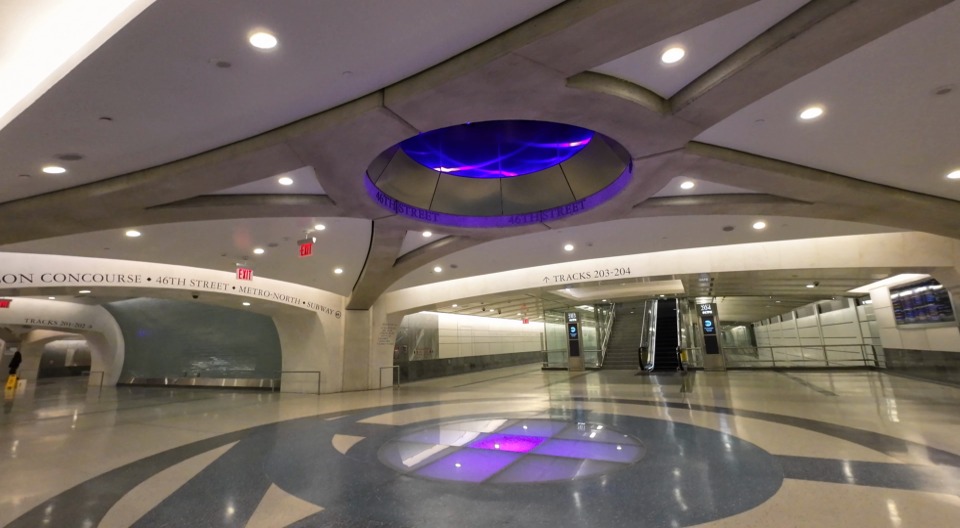} & \qualpanel{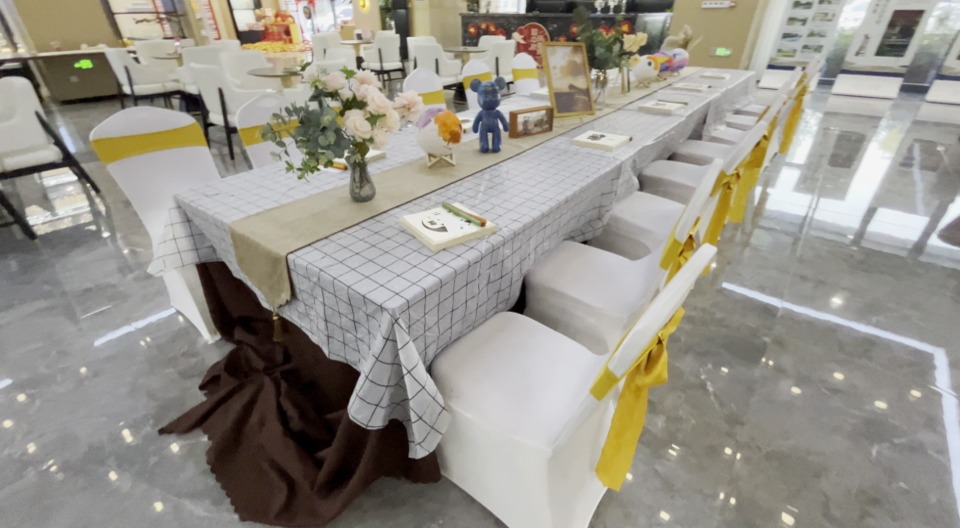} & \qualpanel{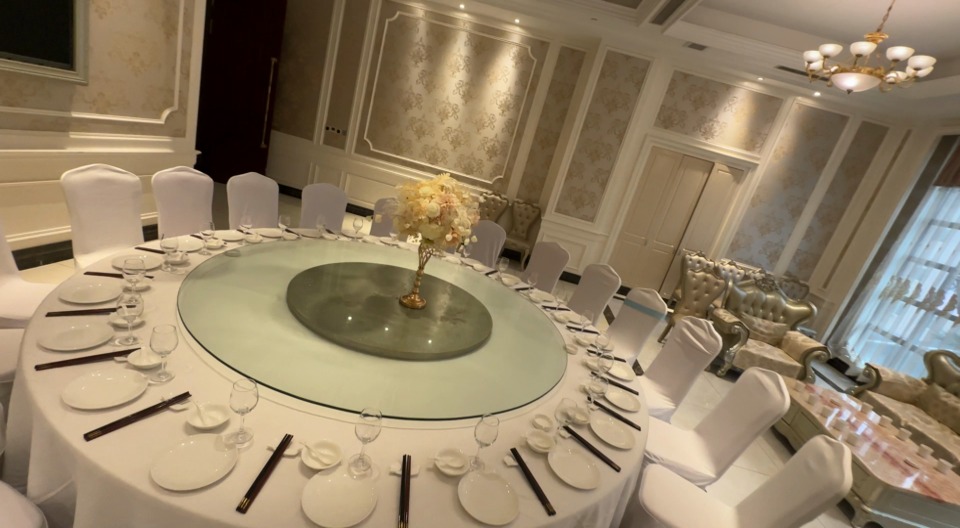} & \qualpanel{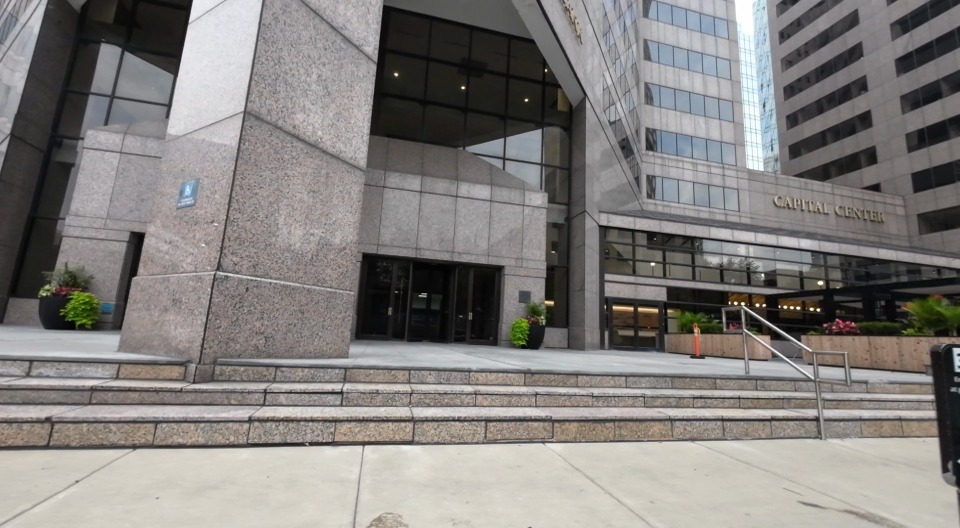} & \qualpanel{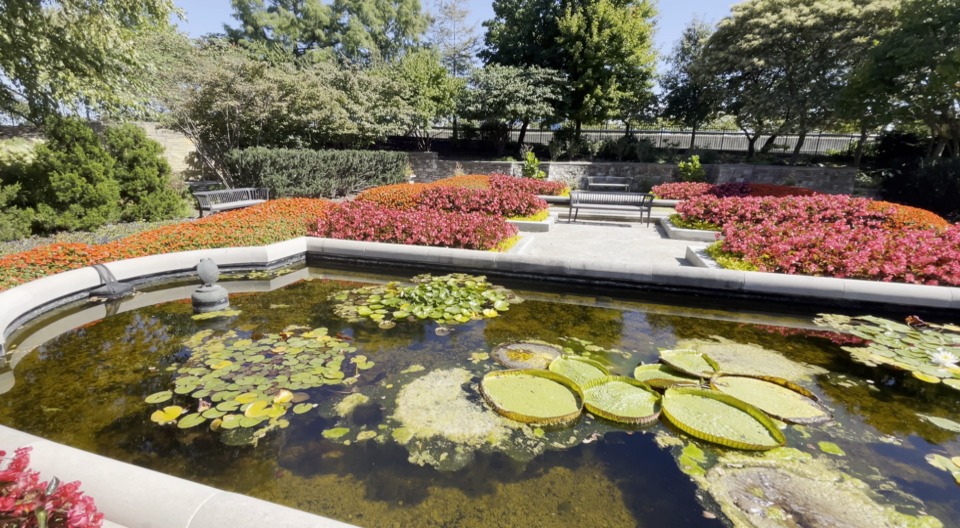} \\
  \end{tabular}}
  \caption{\textbf{Qualitative comparison on DL3DV Hard.} All methods use the same source depth. PE-Field and ViewCrafter tend to distort the geometry when the warp covers little of the target. GEN3C hallucinates the ceiling (first) and renders the pillar at a wrong position (fourth). MetaView shifts the table to the left (second), shrinks the round table so that its near edge enters the view (third), and drops the flowers in the lower left that the warp covers (last). WaveView keeps the layout of the warp and completes what the warp misses.}
  \label{fig:qual_pixel}
\end{figure*}

\subsection{Extension to Dynamic NVS}
\label{sec:video}

For dynamic NVS, each frame keeps its own depth, target camera, resampled source, and covariance, as in \cref{sec:cov}.
The phase injection of \cref{sec:phase} is applied independently to each latent frame, and the frames grouped into one latent frame share one mask $\mathbf{M}_l$.
For a fisheye camera, the source rays and the projection into the target follow its calibrated model, and the epipolar direction of \cref{sec:cov} is mapped to pixels by the Jacobian of the projection.

\section{Experiments}
\label{sec:exp}

\subsection{Setup}
\label{sec:setup}

\paragraph{Benchmarks.}
We evaluate static novel view synthesis on DL3DV~\cite{ling2024dl3dv} following the protocol of MetaView~\cite{cai2026metaview}, and dynamic novel view synthesis on AnyViewBench~\cite{vanhoorick2026anyview}.

\paragraph{Metrics.}
We report PSNR, SSIM~\cite{wang2004ssim}, and LPIPS~\cite{zhang2018lpips}.
For static NVS, we also compute them on the covisible region between target and source, as in DyCheck~\cite{gao2022dycheck} and MegaScenes~\cite{tung2024megascenes}, since only there is the target determined by the source.
This region is found by matching the target with the source using UFM~\cite{zhang2025ufm}, independently of the estimated depth and the warp.
DMD~\cite{cai2026metaview} and further protocol details are in the supplementary material.

\paragraph{Baselines.}
For static NVS, Voyager~\cite{huang2025voyager}, GEN3C~\cite{ren2025gen3c}, and ViewCrafter~\cite{yu2025viewcrafter} complete a reprojection of the source with a video model.
PE-Field~\cite{bai2025pefield} and MetaView~\cite{cai2026metaview} encode the viewpoint change in the positional encoding.
Our static NVS model adapts Qwen-Image-Edit~\cite{wu2025qwenimage} to the structured noise with a LoRA~\cite{hu2022lora} trained on the training set of MetaView.
MetaView shares this generator and our depth estimator Depth Anything 3~\cite{lin2025da3}, but receives the target camera through its positional encoding and a 3D encoder rather than through the initial noise.

For dynamic NVS, we compare with GCD~\cite{vanhoorick2024gcd}, GEN3C, CogNVS~\cite{chen2025cognvs}, TrajCrafter~\cite{yu2025trajcrafter}, and AnyView~\cite{vanhoorick2026anyview}.
Our dynamic NVS model fine-tunes AnyView on the structured noise and keeps its Pl\"ucker conditioning, with training data that follow the distribution of AnyView, restricted to its public datasets: DROID~\cite{khazatsky2024droid} for robotics, ParDom-4D~\cite{vanhoorick2024gcd} and Waymo~\cite{sun2020waymo} for driving, DL3DV~\cite{ling2024dl3dv}, WildRGB-D~\cite{xia2024wildrgbd}, and TartanAir~\cite{wang2020tartanair} for static scenes, and Kubric-5D~\cite{vanhoorick2026anyview} for the remaining domain.
We deliberately train only on datasets close to the pinhole model and leave out Ego-Exo4D~\cite{grauman2024egoexo4d}, whose exocentric frames come from fisheye cameras, showcasing that the noise transfers to a camera model unseen in training.
At test time, we warp its frames with the fisheye calibration of the dataset~\cite{kannala2006generic}.
We were unable to train on LBM~\cite{vanhoorick2026anyview} and Lyft~\cite{houston2020lyft} due to their unavailability and only use evaluation frames republished by AnyView~\cite{vanhoorick2026anyview}.
Details are in the supplementary material.

\begin{figure*}[!t]
  \centering
  \input{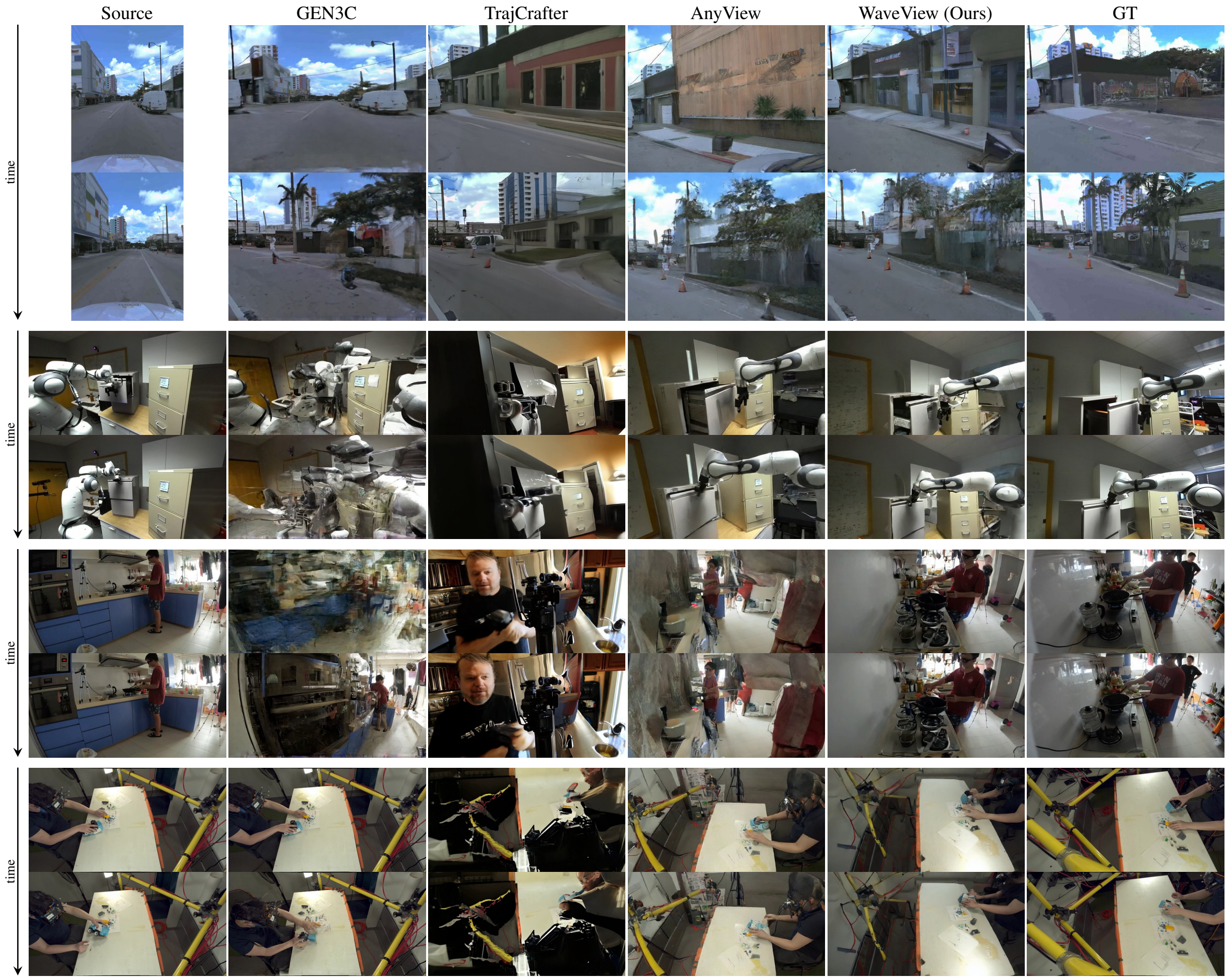}
  \caption{\textbf{Qualitative comparison on the zero-shot split of AnyViewBench}, first and last frame of each target video. GEN3C and TrajCrafter use the same source depth as ours. GEN3C stays at the source viewpoint (first and last example). TrajCrafter renders views that match neither camera, such as a close-up of a person (third). AnyView's views differ from the ground truth in layout: it misplaces the traffic cones (first) and shifts the robot arm (second). WaveView stays closer to the layout of the ground truth.}
  \label{fig:qual_video}
\end{figure*}
\subsection{Quantitative Results}
\label{sec:quant}

\cref{tab:main} compares the methods on the three bins of DL3DV, where WaveView obtains the best results.
On Hard, it gains 0.84\,dB PSNR, 0.052 SSIM, and 0.030 LPIPS over MetaView on the covisible region, and 0.59\,dB, 0.033, and 0.017 on the whole image.
On the covisible region of Easy and Medium, WaveView gains 1.36 and 0.70\,dB PSNR over MetaView, and 0.053 and 0.039 SSIM over GEN3C and ViewCrafter, the strongest baselines in SSIM on these bins.

On the zero-shot split of AnyViewBench (\cref{tab:anyview}), WaveView has the best average in every metric.
It improves AnyView, which shares its generator, from 11.58 to 12.35\,dB PSNR, from 0.331 to 0.372 SSIM, and from 0.621 to 0.587 LPIPS, and it improves all three metrics on each of the five datasets, by up to 1.57\,dB PSNR on Argoverse.
This includes Ego-Exo4D, which is absent from our training data.
On the in-distribution split, all baselines but AnyView are far behind, and WaveView matches AnyView on average, with 17.56 against 17.61\,dB PSNR and slightly better SSIM and LPIPS.
It improves all three metrics on DROID, LBM, Kubric-5D, and ParDom-4D, and trails AnyView on Lyft and Ego-Exo4D, which appear in AnyView's training data but not in ours.
To separate the effect of the structured noise from that of additional training, we also fine-tune AnyView with Gaussian noise using the same training settings.
With Gaussian noise, the fine-tuned model reaches 11.35\,dB PSNR, 0.327 SSIM, and 0.641 LPIPS on the zero-shot split and 17.40\,dB, 0.518, and 0.453 on the in-distribution split, against 12.35\,dB, 0.372, and 0.587 and 17.56\,dB, 0.530, and 0.440 for WaveView.
This indicates that the improvement is not explained by additional fine-tuning alone.

\paragraph{Runtime.}
\label{sec:runtime}
On one B200 GPU, WaveView takes 17.9\,s per view, against 22.0\,s for PE-Field, 38.0\,s for MetaView, 64.1\,s for ViewCrafter, 168.7\,s for GEN3C, and 474.7\,s for Voyager.
MetaView and WaveView both sample 40 steps with a guidance scale of 4 at $960\times528$; their runtime difference comes from the fusion of 3D features inside the denoiser of MetaView at every step.
Depth estimation and the construction of the noise add only 0.46\,s (see the supplementary material).

\subsection{Qualitative Results}
\label{sec:qual}

\cref{fig:qual_pixel} compares the methods on DL3DV Hard.
Explicit warp methods typically fail to preserve the geometry when the source overlaps little with the target view.
MetaView receives the same metric depth through its positional embedding, yet its views are often plausible but deviate from the layout that the warp provides: it shifts or shrinks the table and drops content that the warp covers.
WaveView injects the geometric prior into the phase of the initial noise, so its layout tends to follow the warp where the depth uncertainty is low, while the generator has more freedom where it is high and in disoccluded regions.

\cref{fig:qual_video} shows qualitative results for dynamic NVS.
Where the reprojection covers little of the target, GEN3C keeps the viewpoint of the source and TrajCrafter hallucinates content that appears in neither view.
AnyView receives the target camera only through Pl\"ucker embeddings.
Its outputs are plausible but often deviate from the layout of the ground truth.
With the same generator, WaveView stays closer to it, consistent with its higher PSNR and SSIM (\cref{tab:anyview}).

\subsection{Ablation Study}
\label{sec:ablation}

\begin{figure}[t]
  \centering
  \tikzset{
    wvline/.style={color=wvblue, mark=*, mark size=1.2pt, line width=0.7pt},
    wvrot/.style={color=wvgreen, mark=triangle*, mark size=1.8pt, line width=0.7pt},
    wviso/.style={color=wvorange, mark=diamond*, mark size=1.6pt, line width=0.7pt},
    wvmeta/.style={color=black!55, dashed, line width=0.6pt, mark=none}}
  \begin{tikzpicture}
  \begin{groupplot}[
      group style={group size=3 by 2, horizontal sep=8mm, vertical sep=10mm},
      width=0.43\linewidth, height=0.36\linewidth,
      title style={font=\scriptsize, yshift=-1mm, align=center},
      xlabel style={font=\scriptsize, yshift=1mm}, tick label style={font=\tiny},
      yticklabel style={/pgf/number format/fixed, /pgf/number format/precision=2},
      ymajorgrids, grid style={gray!25}, axis line style={gray!60}, tick style={gray!60},
      enlarge x limits=0.12,
    ]
    \nextgroupplot[xtick={0,0.03,0.06,0.09}, xticklabels={0,.03,.06,.09}, title={Covisible region\\PSNR\,(dB)$\uparrow$},
        legend to name=ablationlegend, legend columns=4,
        legend style={font=\scriptsize, draw=none, column sep=6pt, /tikz/every even column/.append style={column sep=2pt}}]
      \addlegendimage{wvline}\addlegendentry{Ours}
      \addlegendimage{wviso}\addlegendentry{Isotropic}
      \addlegendimage{wvrot}\addlegendentry{Rotated $90^\circ$}
      \addlegendimage{wvmeta}\addlegendentry{Best baseline}
      \addplot[wvmeta, forget plot] coordinates {(0,13.9817) (0.09,13.9817)};
      \addplot[wviso, forget plot] coordinates {(0,14.5593) (0.03,14.5771) (0.06,14.4091) (0.09,14.2197)};
      \addplot[wvline, forget plot] coordinates {(0,14.6422) (0.03,14.8200) (0.06,14.8336) (0.09,14.6817)};
      \addplot[wvrot, forget plot] coordinates {(0,14.6422) (0.03,14.3205) (0.06,14.0557) (0.09,13.7899)};
    \nextgroupplot[xtick={0,0.03,0.06,0.09}, xticklabels={0,.03,.06,.09}, title={Covisible region\\SSIM$\uparrow$}, xlabel={$\rho$ (depth uncertainty) $\uparrow$: more phase released along the epipolar line}]
      \addplot[wvmeta] coordinates {(0,0.3665) (0.09,0.3665)};
      \addplot[wviso] coordinates {(0,0.3913) (0.03,0.3871) (0.06,0.3738) (0.09,0.3614)};
      \addplot[wvline] coordinates {(0,0.3958) (0.03,0.4011) (0.06,0.3967) (0.09,0.3878)};
      \addplot[wvrot] coordinates {(0,0.3958) (0.03,0.3766) (0.06,0.3622) (0.09,0.3513)};
    \nextgroupplot[xtick={0,0.03,0.06,0.09}, xticklabels={0,.03,.06,.09}, title={Covisible region\\LPIPS$\downarrow$}]
      \addplot[wvmeta] coordinates {(0,0.5011) (0.09,0.5011)};
      \addplot[wviso] coordinates {(0,0.4793) (0.03,0.4779) (0.06,0.4925) (0.09,0.5082)};
      \addplot[wvline] coordinates {(0,0.4818) (0.03,0.4714) (0.06,0.4761) (0.09,0.4871)};
      \addplot[wvrot] coordinates {(0,0.4818) (0.03,0.4954) (0.06,0.5135) (0.09,0.5286)};
    \nextgroupplot[xtick={0.3,0.6,0.9,1.2}, xticklabels={.3,.6,.9,1.2}]
      \addplot[wvmeta] coordinates {(0.3,13.9817) (1.2,13.9817)};
      \addplot[wviso] coordinates {(0.3,14.5565) (0.6,14.5771) (0.9,14.4968) (1.2,14.4221)};
      \addplot[wvline] coordinates {(0.3,14.7921) (0.6,14.8200) (0.9,14.6169) (1.2,14.4486)};
      \addplot[wvrot] coordinates {(0.3,14.3172) (0.6,14.3205) (0.9,14.2348) (1.2,14.1263)};
    \nextgroupplot[xtick={0.3,0.6,0.9,1.2}, xticklabels={.3,.6,.9,1.2}, xlabel={$\gamma$ (isotropic floor) $\uparrow$: more phase released in every direction}]
      \addplot[wvmeta] coordinates {(0.3,0.3665) (1.2,0.3665)};
      \addplot[wviso] coordinates {(0.3,0.3884) (0.6,0.3871) (0.9,0.3781) (1.2,0.3667)};
      \addplot[wvline] coordinates {(0.3,0.4027) (0.6,0.4011) (0.9,0.3848) (1.2,0.3666)};
      \addplot[wvrot] coordinates {(0.3,0.3778) (0.6,0.3766) (0.9,0.3666) (1.2,0.3540)};
    \nextgroupplot[xtick={0.3,0.6,0.9,1.2}, xticklabels={.3,.6,.9,1.2}]
      \addplot[wvmeta] coordinates {(0.3,0.5011) (1.2,0.5011)};
      \addplot[wviso] coordinates {(0.3,0.4774) (0.6,0.4779) (0.9,0.4815) (1.2,0.4900)};
      \addplot[wvline] coordinates {(0.3,0.4741) (0.6,0.4714) (0.9,0.4824) (1.2,0.4951)};
      \addplot[wvrot] coordinates {(0.3,0.4964) (0.6,0.4954) (0.9,0.5019) (1.2,0.5113)};
  \end{groupplot}
  \end{tikzpicture}\\[1pt]
  \ref*{ablationlegend}
  \caption{\textbf{Ablation on DL3DV Hard} (covisible region). Isotropic: trained with $\mathbf{\Sigma}$ replaced by $\frac12\operatorname{tr}(\mathbf{\Sigma})\mathbf{I}$. Rotated $90^\circ$: our model with the depth covariance rotated at inference. Best baseline: best of all baselines per metric.}
  \label{fig:ablation}
\end{figure}
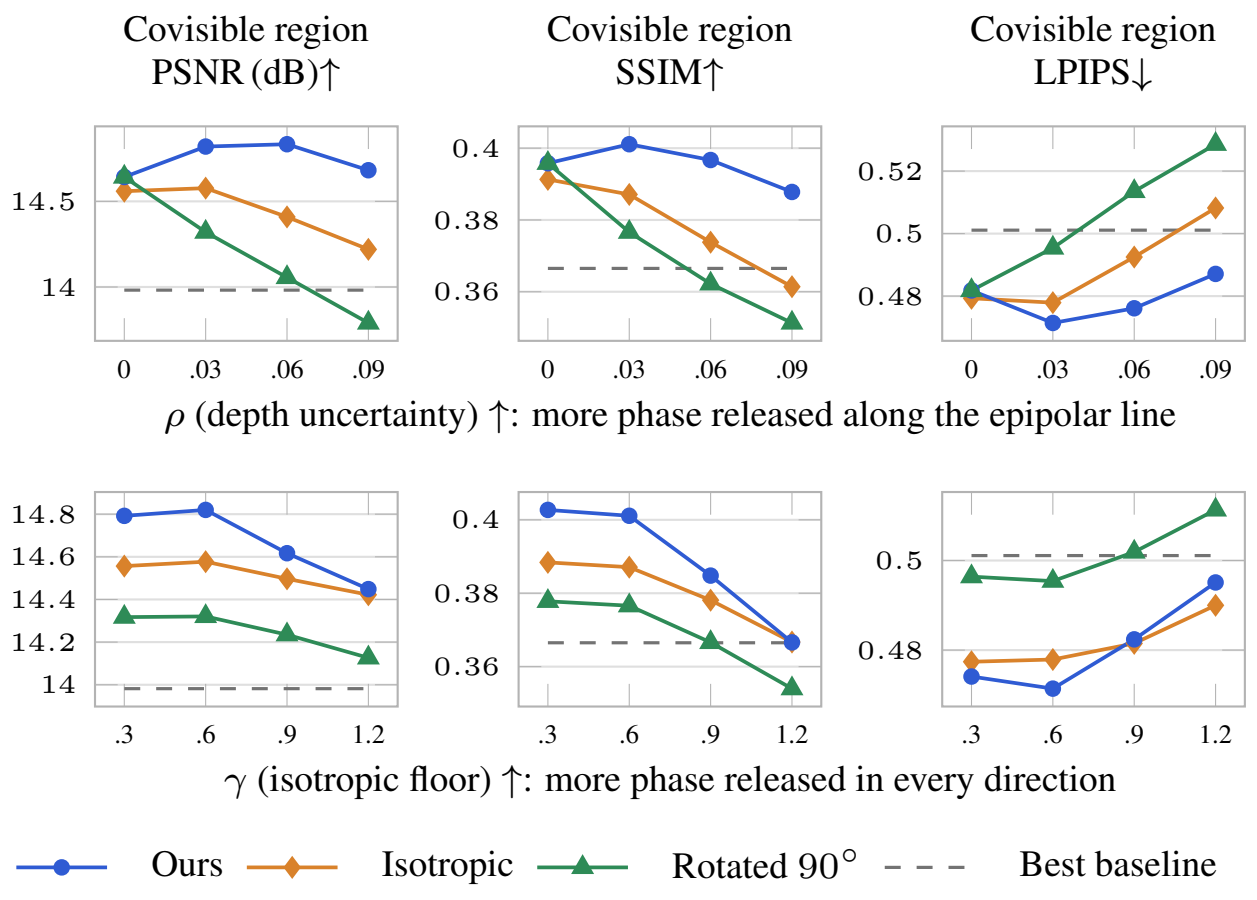

\begin{figure}[t]
  \centering
  {\setlength{\tabcolsep}{0.5pt}\renewcommand{\arraystretch}{0.6}%
  \newcommand{\aqrow}[1]{\raisebox{0.065\linewidth}{\rotatebox[origin=c]{90}{\scriptsize #1}}\hspace{1pt}}%
  \newcommand{\aqimg}[1]{\includegraphics[width=0.237\linewidth]{figures/aniso_qual/#1}}%
  \begin{tabular}{@{}ccccc@{}}
    & {\scriptsize Source} & {\scriptsize Variant} & {\scriptsize Ours} & {\scriptsize Ground truth} \\
    \aqrow{Isotropic} & \aqimg{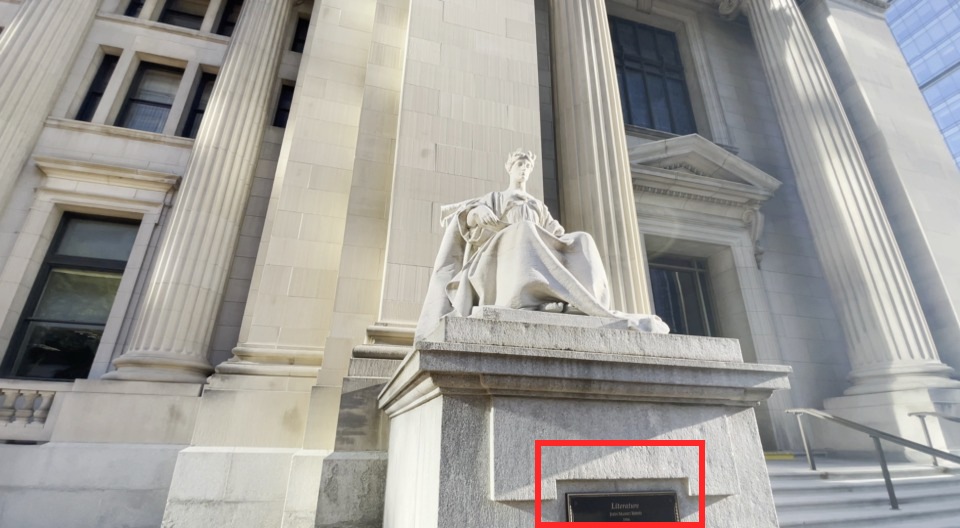} & \aqimg{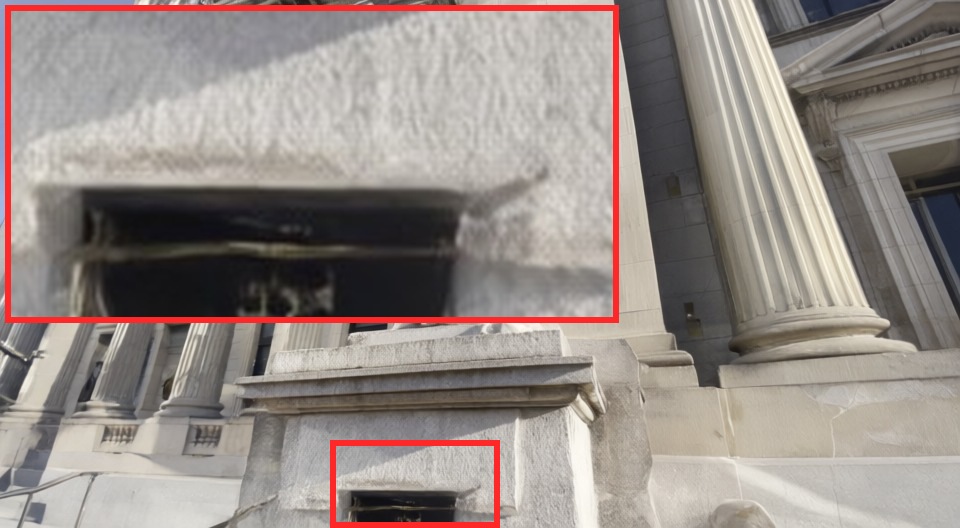} & \aqimg{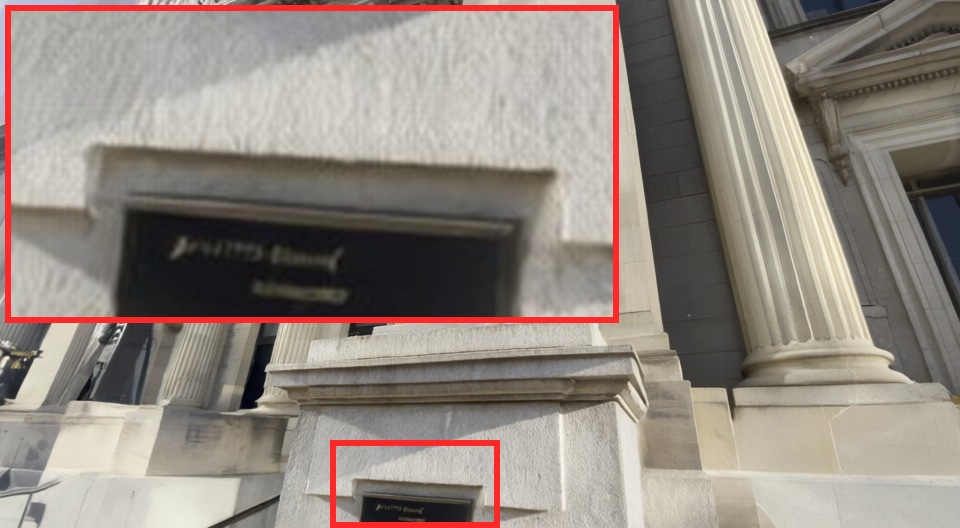} & \aqimg{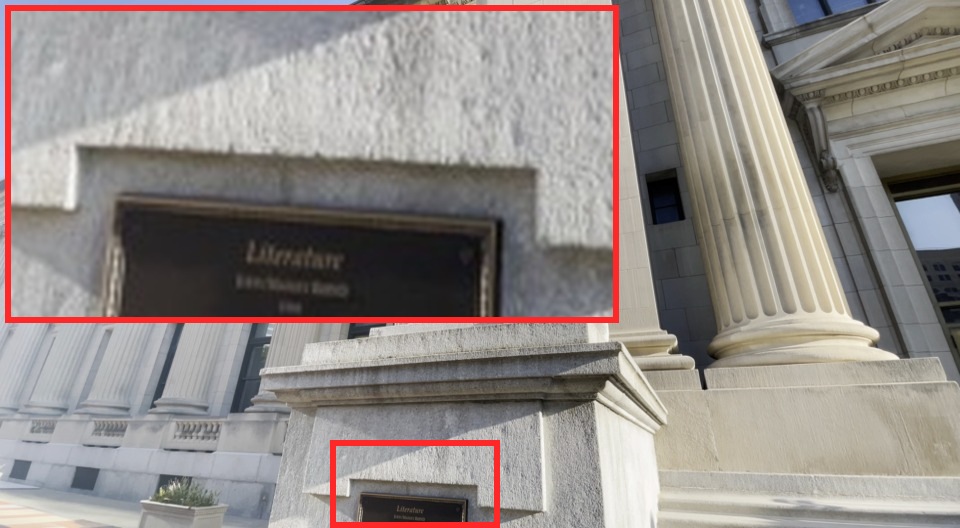} \\
    \aqrow{Low-pass} & \aqimg{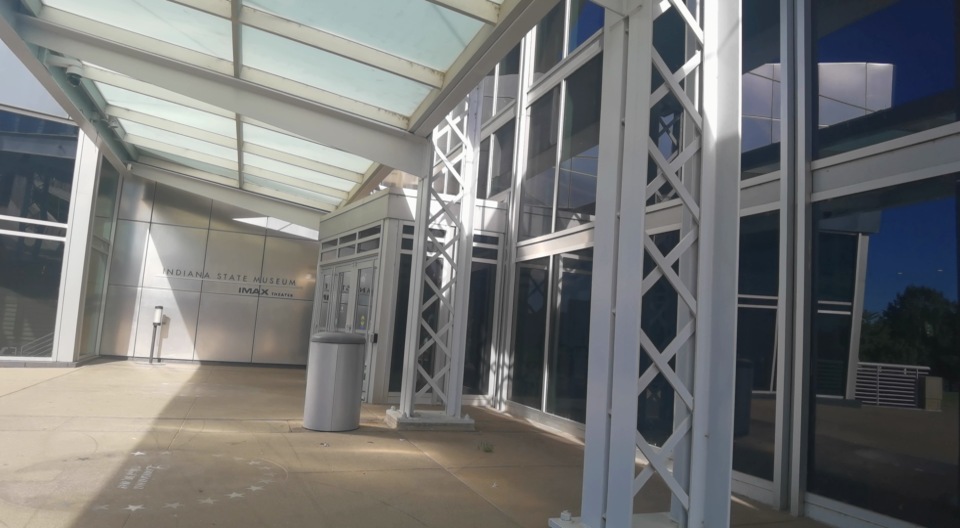} & \aqimg{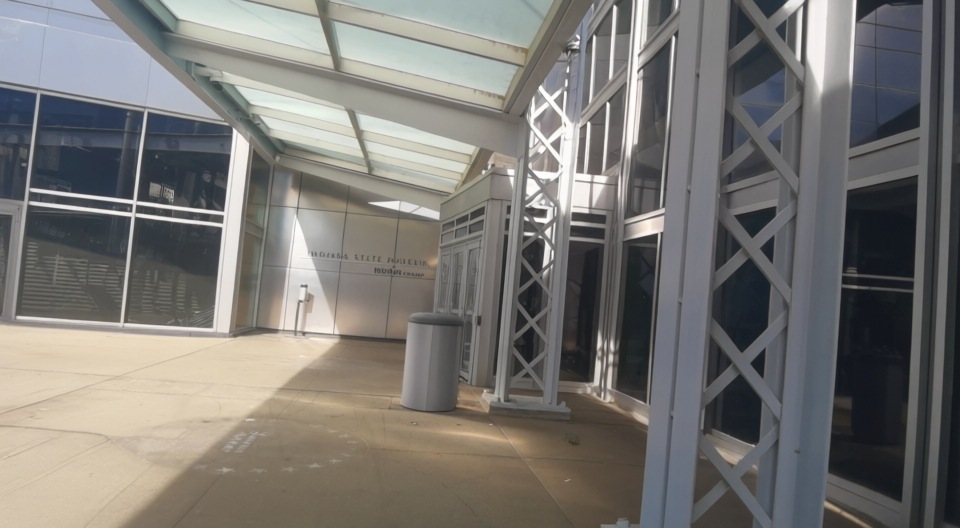} & \aqimg{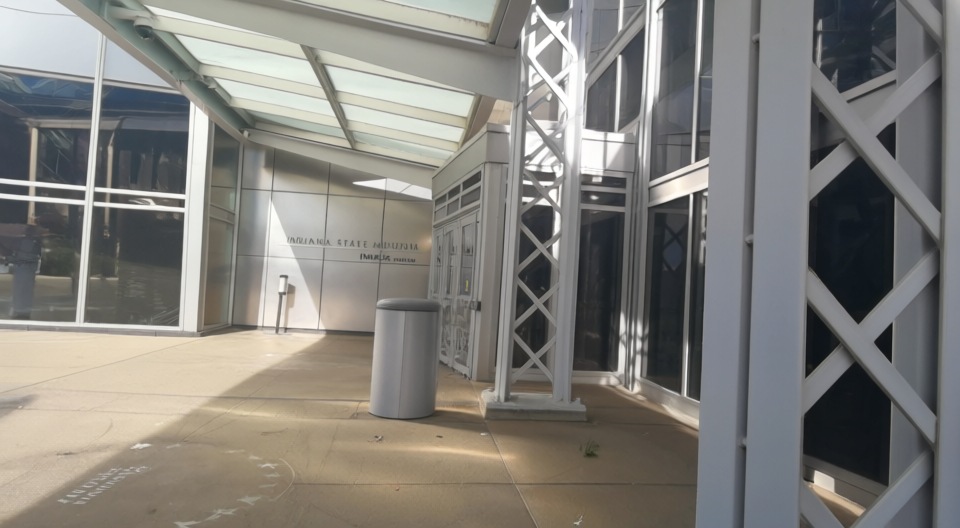} & \aqimg{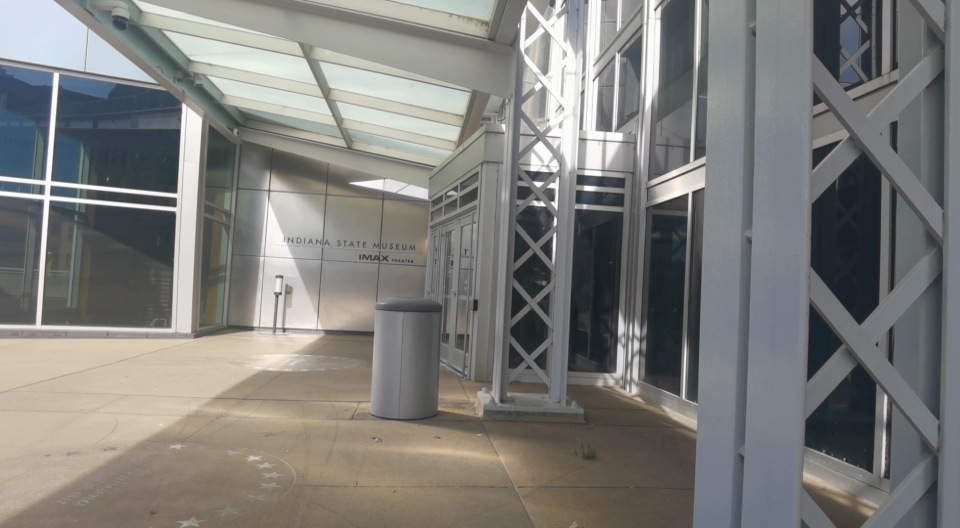} \\
  \end{tabular}}
  \caption{\textbf{Variants of the phase injection versus ours} on DL3DV Hard. Top: with the same seed and warp, the isotropic variant breaks the frame of the plaque (enlarged boxes). Bottom: the low-pass variant shifts the truss and the column off their ground-truth positions, while ours keeps them.}
  \label{fig:aniso_qual}
\end{figure}

\paragraph{Anisotropic vs.\ isotropic.}
\cref{fig:ablation} studies the direction of $\mathbf{\Sigma}$ and the effects of $\rho$ and $\gamma$ on DL3DV Hard; the default $\rho=0.03$ is the error of the depth estimator measured on ground-truth depth (see the supplementary material), and the sweeps vary $\rho$ and $\gamma$ at test time.
We train an isotropic variant with the same recipe, replacing $\mathbf{\Sigma}$ by $\frac12\operatorname{tr}(\mathbf{\Sigma})\mathbf{I}$, and rotate $\mathbf{\Sigma}$ of our model by $90^\circ$ at inference, releasing the phase across the epipolar line; both keep $\operatorname{tr}(\mathbf{\Sigma})$ and change only its direction.
At the default setting ($\rho=0.03$, $\gamma=0.6$), our model is better than the isotropic variant by 0.24\,dB PSNR and 0.014 SSIM, and than the rotated one by 0.50\,dB and 0.025.
In PSNR, ours stays above the isotropic variant and the isotropic variant above the rotated one at every nonzero $\rho$ and every $\gamma$, and the gaps of ours to the isotropic and the rotated variant grow with $\rho$, from 0.24 to 0.46\,dB and from 0.50 to 0.89\,dB as $\rho$ increases from 0.03 to 0.09: with the same total uncertainty, the direction in which the phase is released matters.
\cref{fig:aniso_qual} (top) shows one pair: with the same seed and warp, the isotropic variant breaks the frame of the plaque that the source observes, while our model keeps it.

\paragraph{Depth uncertainty $\rho$.}
Setting $\rho=0$ lowers PSNR by 0.18\,dB from the default and worsens SSIM and LPIPS, since $\mathbf{M}_l$ then keeps the phase also where the warp is misplaced along the epipolar line.
At $\rho=0.06$, PSNR is similar but SSIM and LPIPS are slightly worse, and at $\rho=0.09$, all three metrics worsen, as too much of the geometric prior is released.

\paragraph{Isotropic floor $\gamma$.}
The results change little between $\gamma=0.3$ and $0.6$.
A larger $\gamma$ releases the phase in every direction: at $\gamma=1.2$, PSNR drops by 0.37\,dB, and the advantage over the isotropic variant shrinks to 0.03\,dB, as the isotropic floor then dominates the directional term.

\paragraph{Low-pass injection.}
Trained with the same recipe but taking only $\mathbf{L}$ from the warp and every band phase from the noise, similar to WAVE~\cite{park2025wave}, the model reaches 12.85\,dB PSNR, 0.307 SSIM, and 0.578 LPIPS on the covisible region of Hard, 1.97\,dB below ours and below MetaView, suggesting that much of the layout is conveyed by the phase of the bands (\cref{fig:aniso_qual}, bottom).

\section{Conclusion}
\label{sec:conclusion}

We presented WaveView, which injects a geometric prior of the novel view into the phase of a pretrained generator's initial noise.
WaveView releases phase along the epipolar line as depth uncertainty grows, preserves it across the line, and leaves disoccluded regions as noise.
This gives the generator a directional geometric prior with room to correct uncertain content, with negligible inference overhead.
On DL3DV, WaveView outperforms all compared methods while running faster, and the isotropic and low-pass comparisons support anisotropic phase injection.
For dynamic NVS, the same construction improves AnyView on all three average metrics of the zero-shot split of AnyViewBench.
\paragraph{Limitations.}
$\mathbf{\Sigma}$ models only a local error of the estimated depth along each ray, not a wrong global scale, an inaccurate camera, or a surface that the warp hides but the target shows.
Where the source has no pixel, the content is left entirely to the learned distribution of the generator.

{
    \small
    \bibliographystyle{ieeenat_fullname}
    \bibliography{main}
}

\end{document}